\documentclass[review]{acmtog}
\usepackage[caption=false,font=footnotesize]{subfig}
\usepackage{amsmath,amsfonts,amssymb}
\usepackage{graphicx}
\usepackage{prettyref}
\usepackage{mathrsfs}
\usepackage{wrapfig}
\usepackage{multirow}
\usepackage{xcolor}
\usepackage{booktabs}
\usepackage{algorithm,algpseudocode}
\newrefformat{Fig}{Figure~\ref{#1}}
\newrefformat{fig}{figure~\ref{#1}}
\newrefformat{par}{section~\ref{#1}}
\newrefformat{appen}{appendix~\ref{#1}}
\newrefformat{sec}{section~\ref{#1}}
\newrefformat{sub}{section~\ref{#1}}
\newrefformat{table}{table~\ref{#1}}
\newrefformat{alg}{algorithm~\ref{#1}}
\newrefformat{Def}{definition~\ref{#1}}
\newrefformat{Thm}{theorem~\ref{#1}}
\newrefformat{step}{step~\ref{#1}}
\newrefformat{eq}{~(\ref{#1})}
\newrefformat{pb}{problem~\ref{#1}}
\def\Eqref Eq:#1:{\eqref{eq:#1}}
\newrefformat{Eq}{Equation~\Eqref#1:}

\newcommand{\E}[1]{\mathbf{#1}}

\usepackage{tikz}
\usetikzlibrary[arrows,backgrounds,decorations.pathmorphing,positioning,fit,petri]
\DeclareMathOperator*{\argmin}{\arg\!\min}
\acmVolume{VV}
\acmNumber{N}
\acmYear{YYYY}
\acmMonth{Month}
\acmArticleNum{XXX}  
\acmdoi{10.1145/XXXXXXX.YYYYYYY}
\begin{document}
\title{Detailed Garment Recovery from a Single-View Image}
\author{Shan Yang, Tanya Amert, Zherong Pan, Ke Wang, Licheng Yu, Tamara Berg {\upshape and} Ming C. Lin 
\affil{University of North Carolina at Chapel Hill}}




%\acmformat{Pamplona, V. F., Oliveira, M. M., and Baranoski, G. V. G. 2009. Photorealistic models for pupil light
%reflex and iridal pattern deformation.  {ACM Trans. Graph.} 28, 4, Article 106 (August 2009), 11 pages.\newline  DOI $=$
%10.1145/1559755.1559763\newline http://doi.acm.org/10.1145/1559755.1559763}

\maketitle
\begin{bottomstuff} 
Authors' addresses: land and/or email addresses.
\end{bottomstuff}
\begin{abstract}
%Capturing highly detailed human body parts have recently become an active area of research. In this work, we address the problem of joint capture of both human body shape, pose and highly detailed clothing mesh. Unlike previous methods relying on multiview imaging, video sequence or RGB-D camera, our algorithm takes as input only a single-view internet image. To address such a highly ill-posed problem, we first optimize for a coarse human pose, using a reduced human model learned from a dressed human body database. Then system can then recover both the human body shape, pose and highly detailed cloth by a heuristic optimization taking into account the imaging alignment, wrinkle consistency and physical validity.
 Most recent garment capturing techniques rely on acquiring multiple views of clothing, which may not always be readily available, especially in the case of pre-existing photographs from the web. As an alternative, we propose a method that is able to compute a 3D model of a human body and its outfit from a single 
photograph % of a person wearing clothing 
with little human interaction. Our algorithm is not only able to capture the global shape and geometry of the clothing, it can also extract small but important details of cloth, such as occluded wrinkles and folds. Unlike previous methods using full 
3D information (i.e. depth, multi-view images, or sampled 3D geometry), our 
approach achieves detailed garment recovery from a single-view image by using 
statistical, geometric, and physical priors and a combination of parameter estimation, semantic parsing, shape recovery, and physics-based cloth simulation. 
We demonstrate the effectiveness of our algorithm by re-purposing the 
reconstructed garments for  virtual try-on and 
garment transfer applications and for cloth animation for digital characters.
\end{abstract}
%%% The first section of your paper. 
\newenvironment{revision}
{\color{black}}

\section{Introduction}

Retail is a multi-trillion dollar business worldwide, with the global fashion industry
valued at \$3 trillion~\cite{fashion_stat}.  Approximately \$1.6 trillion of retail 
purchasing in 2015 was done via online e-commerce sales, with growth rates in the double digits~\cite{ecommerce_stat}.
% {https://www.internetretailer.com/2015/07/29/global-e-commerce-set-grow-25-2015}. 
Thus, enabling better online apparel shopping experiences has the potential for enormous economic impact. Given the worldwide demand for fashion and the obvious impact of this demand on the apparel industry, technology-based solutions have recently been proposed, a few of which are already in use by commercial vendors. For example, there are several computer-aided design (CAD) software systems developed specifically for the apparel industry. The apparel CAD industry has focused
predominantly on sized cloth-pattern development, pattern-design CAD software, 3D draping preview, and automatic 
generation of 2D clothing patterns from 3D body scanners or other measurement devices. 
Some of the leading apparel CAD companies include Gerber Technology, Lectra, Optitex,
Assyst, StyleCAD, Marvelous Designer, Clo-3D and Avametric, etc. Unfortunately, developing these systems requires careful and lengthy design by an apparel expert. 

More recently, there have been efforts to develop virtual try-on systems, such as triMirror, that allow users to visualize what a garment might look like on themselves before purchasing. These methods 
enable 3D visualization of various garments, fast animation of dynamic cloth, and a quick preview of 
how the cloth drapes on avatars as they move around. However, the capabilities of these new 
systems are limited, especially in terms of ease-of-use and applicability. Many of the virtual try-on systems use simple, 
fast image-based or texture-based techniques for a fixed number of avatar poses. 
%They do not perform a complete simulation of how the fabric bends, wrinkles, or changes its physical appearance when the virtual human bends, stretches, or changes his/her pose in various activities. 
They do not typically account for the effects on fabric materials of different conditions 
(e.g. changing weather, varying poses, 
weight fluctuation, etc). Furthermore, almost all of these virtual try-on systems 
assume either that the user selects one of a pre-defined set of avatars or 
that accurate measurements of their own bodies have been captured via 3D scans.

% Capturing high-resolution, functional models for human bodies has been made possible over the 
% last few years by incorporating body-part dependent priors for hair~\cite{chai2012single}, 
% face~\cite{cao20133d}, skin~\cite{Nagano:2015:SMD:2809654.2766894}, or eyeball~\cite{berard2014high}. 
% However, some non-body parts can also drastically increase the visual realism of a human body model, 
% of which the most common is clothing.
\begin{figure*}[th]
\centerline{
\includegraphics[width=1.0\textwidth]{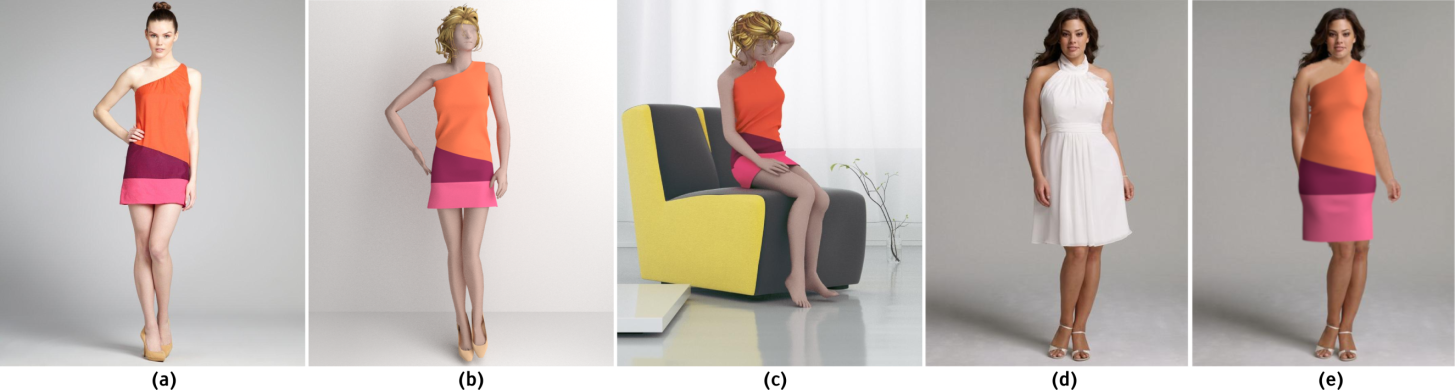}
}
\caption{{\bf Garment recovery and re-purposing results.} From left to right, we show an example of (a) the original image {\em \protect\cite{dress1}~$\copyright$}, (b) the recovered dress and body shape from a single-view image, and
 (c)-(e) the recovered garment on another body of different poses and shapes/sizes 
{\em \protect\cite{garmentTransfer5}~$\copyright$}.}
\label{fig:cover}
\end{figure*}

\begin{figure*}[th]
\centerline{
\includegraphics[width=7.2in]{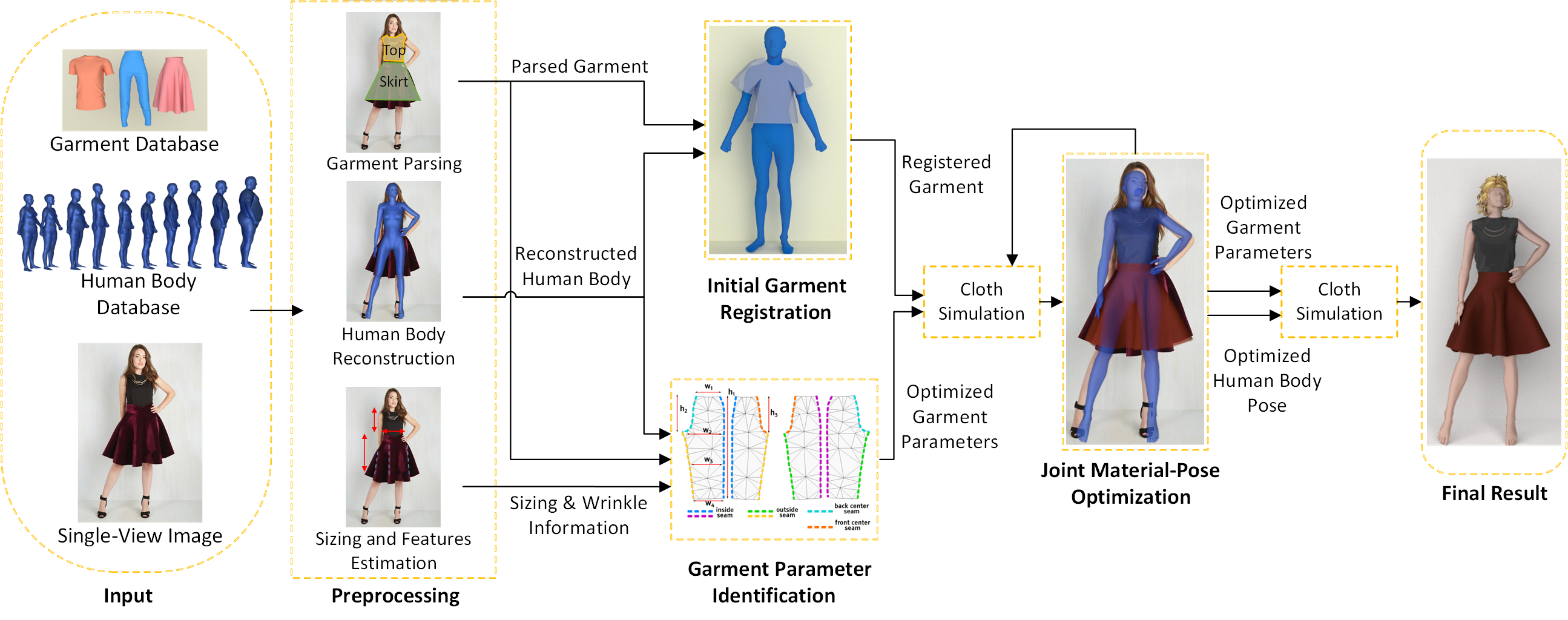}
}
\caption{{\bf The flowchart of our algorithm. } We take a single-view image 
{\em \protect\cite{skirt}~$\copyright$}, a human-body dataset, and a garment-template database as input. 
We preprocess the input data by performing garment parsing, sizing and features estimation, and human-body reconstruction. 
Next, we recover an estimated garment described by the set of garment parameters, including 
fabric material, design pattern parameters, sizing and wrinkle density, as well as the 
registered garment dressed on the reconstructed body.}
\label{methodOverviewFig}
\end{figure*}

In this work, we consider the problem of recovering detailed models of garments from a single-view image. Such a capability enables users to virtually try on garments given only a single photograph of themselves wearing clothing. 
Instead of representing the clothed human as a single mesh~\cite{chen2013deformable,li2012temporally}, we define a separate mesh for a person's clothing, allowing us to model the rich physical interactions between clothing and the human body. This approach also helps capture occluded wrinkles in clothing that are caused by various sources, including garment design that incorporates pleats, cloth material properties that influence the drape of the fabric, and the underlying human body pose and shape. Figure~\ref{fig:cover} illustrates some results generated by our system. In addition to virtual try-on applications, broader impacts in graphics include improving the accuracy of clothing models for animated characters, with the potential to further increase the visual realism of digital human models that already incorporate body-dependent priors for hair~\cite{chai2012single}, face~\cite{cao20133d}, skin~\cite{Nagano:2015:SMD:2809654.2766894}, and eyeballs~\cite{berard2014high}.

With limited input from a single-view image, we constrain the problem's solution space by exploiting three important priors.  The first prior is a statistical human body distribution model constructed from a (naked) human body data set.
This statistical model is used for extracting and matching the human body shape and pose in a given input image. The second prior is a collection of
all sewing patterns of various common garment types, such as skirts, 
pants, shorts, t-shirts, tank tops, and dresses,  
from a database of all garment templates. 
Finally, the third prior is a set of
all possible configurations and dynamical states
of garments governed by their respective constitutive laws and
modeled by a physically-based cloth simulation.  Simulation helps 
provide additional 3D physical constraints lacking in a 2D image.  

Our method proceeds as follows. To construct an accurate body model, the user indicates 14 joint positions on the image and provides a rough sketch outlining the human body silhouette. (This step can also be automated using image processing and body templates for standard unoccluded poses.)  From this information, we use a statistical human model to automatically generate a human body mesh for the image. To estimate the clothing model, we first compute a semantic parse of the garments
in the image to identify and localize depicted clothing items. This semantic segmentation is computed automatically using a data-driven method for clothing recognition~\cite{yamaguchi2013paper}. We then use the semantic parsing to extract garment sizing information, such as waist girth, skirt length and so on, which are then used to map the depicted garments onto the existing garment templates and adjust the sewing patterns based on the extracted parameters. 
We also analyze the segmented garments to identify the location and density of wrinkles and folds in the recovered garments, which are necessary for estimating material properties of the garments for virtual try-on. 

Once we have obtained both the body and clothing models, we perform an image-guided parameter identification process, which optimizes the garment template parameters based on the reconstructed human body and image information. We fit our 3D garment template's surface mesh onto the human body to obtain the initial 3D garment, then jointly optimize the material parameters, the 
body shape, and the pose to obtain the final result. % by incorporating information about the physical properties of cloth.  
The flow chart of the overall process is shown in Fig.~\ref{methodOverviewFig}.  Our main contributions include:
\begin{itemize}
    \item An image-guided garment parameter selection method that makes the generation of virtual garments with diverse styles and sizes a simple and natural task (Section~\ref{garmentParameterSec});
    \item A joint material-pose optimization framework that can reconstruct both body and cloth models with material properties from a single image (Section~\ref{sec:fitting});
    \item Application to virtual try-on and character animation (Section~\ref{sec:results}).
\end{itemize}

% Next, we review some of the related work, followed by a formal definition of the problem statement and assumptions made in our system.  

\section{Related Work}
Our work is built on previous efforts in cloth modeling, human pose\slash shape recovery, garment capture from single-view images, and semantic parsing.

{\bf Cloth Modeling:}
Cloth simulation is a traditional research problem in computer graphics.
Early work on cloth simulation includes~\cite{weil1986synthesis,ng1996computer,baraff1998large,house2000cloth}.  
More recently, a number of methods were proposed to solve the complicated problems presented in cloth simulation, including collision detection~\cite{govindaraju2007fast,tang2009iccd,curtis2008fast}, collision handling, friction handling~\cite{bridson2002robust}, strain limiting~\cite{goldenthal2007efficient,english2008animating,thomaszewski2009continuum,wang2010multi} and remeshing~\cite{narain2012adaptive}. 

Realistic wrinkle simulation is an important problem in realistic cloth modeling.  
Volino and Magnenat-Thalmann~\shortcite{volino1999fast} introduced a geometry-based wrinkle synthesis.
Rohmer et al.~\shortcite{rohmer2010animation} presented a method to augment a coarse cloth mesh with wrinkles. 
Physically based cloth wrinkle simulation depends on an accurate model of the underlying constitutive law;
different bending and stretching energy models for wrinkle simulation have been proposed~\cite{bridson2003simulation}. 

%In this paper, our goal is to recover realistic garments from a single-view image. 
Garment modeling is built upon cloth simulation.
It also needs to take into consideration the design and sewing pattern of the garment. 
Some methods start from the 2D design pattern~\cite{protopsaltou2002body,decaudin2006virtual,berthouzoz2013parsing} or 2D sketches~\cite{turquin2007sketch,robson2011context}. 
Other methods explore garment resizing and transfer from 3D template garments~\cite{wang2005design,meng2012flexible,sumner2004deformation}.  
\begin{revision}
In contrast, our work synthesizes different ideas and extends these methods to process 2D input image and fluidly transfer the results to the simulation of 3D garments. 
We can also edit the 2D sewing patterns with information extracted from a 
single-view image, which can be used to guide the generation of garments 
of various sizes and styles. 
\end{revision}

{\bf Human Pose and Shape Recovering:}
Human pose and shape recovery from a single-view image has been extensively studied in computer vision and computer graphics.  
Taylor~\shortcite{taylor2000reconstruction} presented an articulated-body skeleton recovery algorithm from a single-view image with limited user input. 
Agarwal et al.~\shortcite{agarwal2006recovering} proposed a learning-based method to recover human body poses from monocular images. 
Ye et al.~\shortcite{ye2014real} applied a template-based method for real-time single RGBD image human pose and shape estimation. 
We refer readers to this survey on human motion recovery and analysis~\cite{moeslund2006survey}.

Human pose and shape recovery in computer graphics focus primarily on reconstructing muscle accurately and on watertight 3D human body meshes.  
A realistic 3D human body mesh is the basis for character animation. 
A human body mesh is required for the recovery of clothing with rich details. 
For human body mesh generation, we follow the previous data-driven methods, most of which are PCA based.
These techniques use a set of bases to generate a variety of human bodies of different poses and shapes. 
Seo and Thalmann~\shortcite{seo2003automatic} presented a method to construct human body meshes of different shapes.  
Following this work, Anguelov et al.~\shortcite{anguelov2005scape} introduced the SCAPE model, which can produce human body meshes of different poses and shapes.  
Using the SCAPE model, Balan et al.~\shortcite{balan2007detailed} presented a method to recover detailed human shape and pose from images. 
Hasler et al.~\shortcite{hasler2009statistical} encode both human body shapes and poses using PCA and semantic parameters.   
Building upon these previous models, Zhou et al.~\shortcite{zhou2010parametric} proposed a method to recover the human body pose and shape from a single-view image.  

{\bf Clothing Capturing:} 
In the last decade, many methods have been proposed for capturing clothing from images or videos.   
Methods can be divided into two categories: marker-based and markerless. 
Most marker-based clothing capture methods require the markers to have been pre-printed on the surface of the cloth. 
\begin{revision}
Different kinds of markers have been used for the capture~\cite{scholz2006texture,hasler2006physically,tanie2005high,scholz2005garment,white2007capturing}. 
Markerless methods, which do not require pre-printed clothing markers, can be characterized into several categories of methods: single-view~\cite{zhou2013garment,jeong2015garment}, 
depth camera based~\cite{chen2015garment}; and multi-view methods~\cite{popa2009wrinkling}.
Another method based on cloth contact friction design was proposed by Casati et. al.~\shortcite{casati2016inverse}.
\end{revision}
These methods have some limitations, however, including inability to capture fine garment details and material properties, the loss of the original garment design, and complexity of the capturing process. 
In contrast, our method can retrieve the 2D design pattern with the individual 
measurements obtained directly from a single image. 
Using a joint human pose and clothing optimization method, our algorithm recovers realistic 
garment models with details (e.g. wrinkles and folds) and material properties.  

{\bf Semantic Parsing:} Semantic parsing is a well-studied problem in computer vision, where the goal is to assign a semantic label to every pixel in an image. Most prior work has focused on parsing general scene images~\cite{long_shelhamer_fcn,Farabet13,Pinheiro14}. We work on the somewhat more constrained problem of parsing clothing in an image. To obtain a semantic parse of the clothing depicted in an image, we make use of the data-driven approach by Yamaguchi et al.~\shortcite{yamaguchi2013paper}. This method automatically estimates the human body pose from a 2D image, extracts a visual representation of the clothing the person is wearing, and then visually matches the outfit to a large database of clothing items to compute a 
clothing parse of the query image.

\section{Problem Statement and Assumptions}

In this section, we give the formal definition of the problem. The input to our system is an RGB image $\Omega$. We assume the image is comprised of three parts: the background region $\Omega_b$, the foreground naked human body parts $\Omega_h$ and the foreground garment $\Omega_g$, where 
$\Omega = \Omega_b \cup \Omega_h \cup \Omega_g$.
In addition, we assume that both the human body and the garment are in a statically stable physical state. Although this assumption precludes images capturing a fast moving human, it provides a crucial assumption for our joint optimization algorithm.

\begin{revision}
\vspace*{0.5em}
\noindent
{\bf Problem: } Given $\Omega_g$, $\Omega_h$, how to recover 

\vspace*{0.5em}
\noindent
-- the garment described by a set of parameters $<\mathcal{C},\mathcal{G},\E{U},\E{V}>$, \\

\vspace*{-0.5em}
\noindent
-- along with a set of parameters $<\boldsymbol\theta,\E{z}>$ that encode human body pose and shape obtained from the image.  
%\vspace*{-0.5em}
%\noindent
%-- and a set of parameters $<\mathcal{L},\mathcal{P}>$ defining the 2D sizing parameters and wrinkle density of the reference garment. 
\end{revision}

\begin{figure*}[t]
\vspace{-10mm}
\hspace{-6mm}
\begin{center}
\subfloat[]{
       \includegraphics[width =0.15\textwidth] {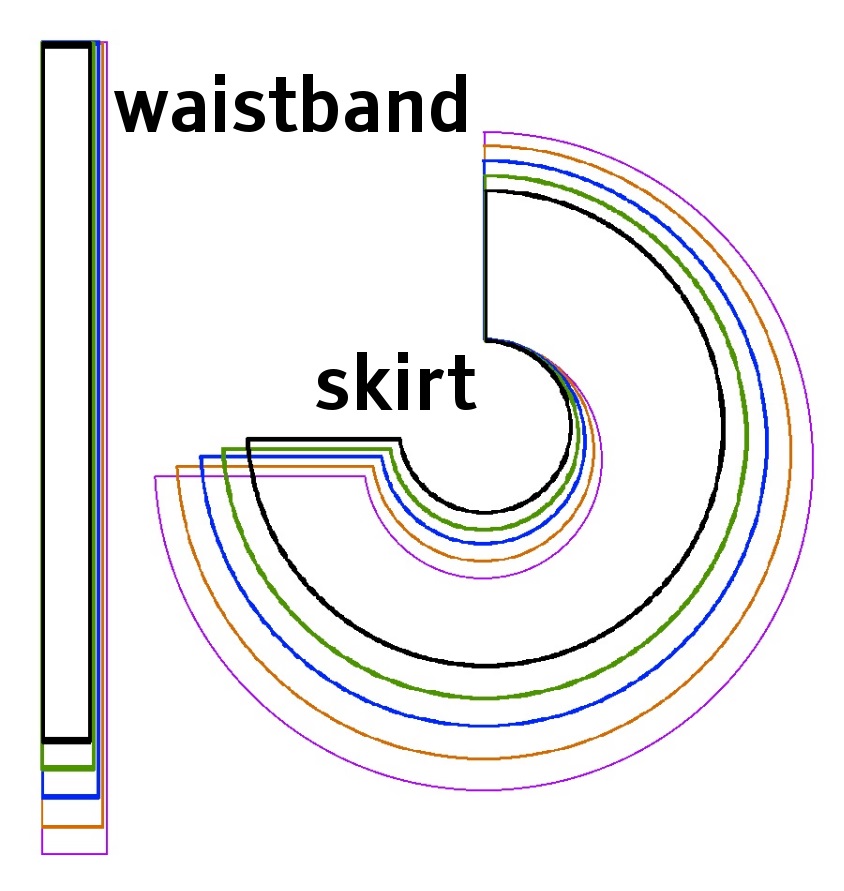}
          \label{classicSkirtSewingPatternFig}
      }
\subfloat[]{
       \includegraphics[width =0.17\textwidth] {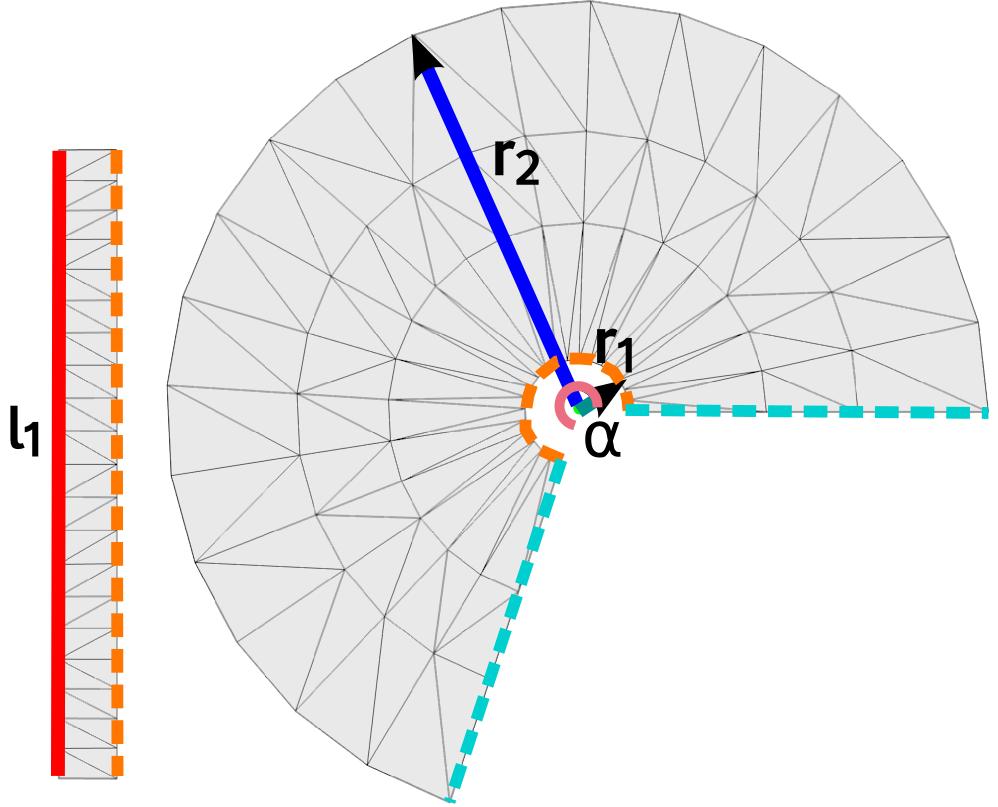}
          \label{skirtMSFig}
      }
\subfloat[]{
       \includegraphics[scale =0.21] {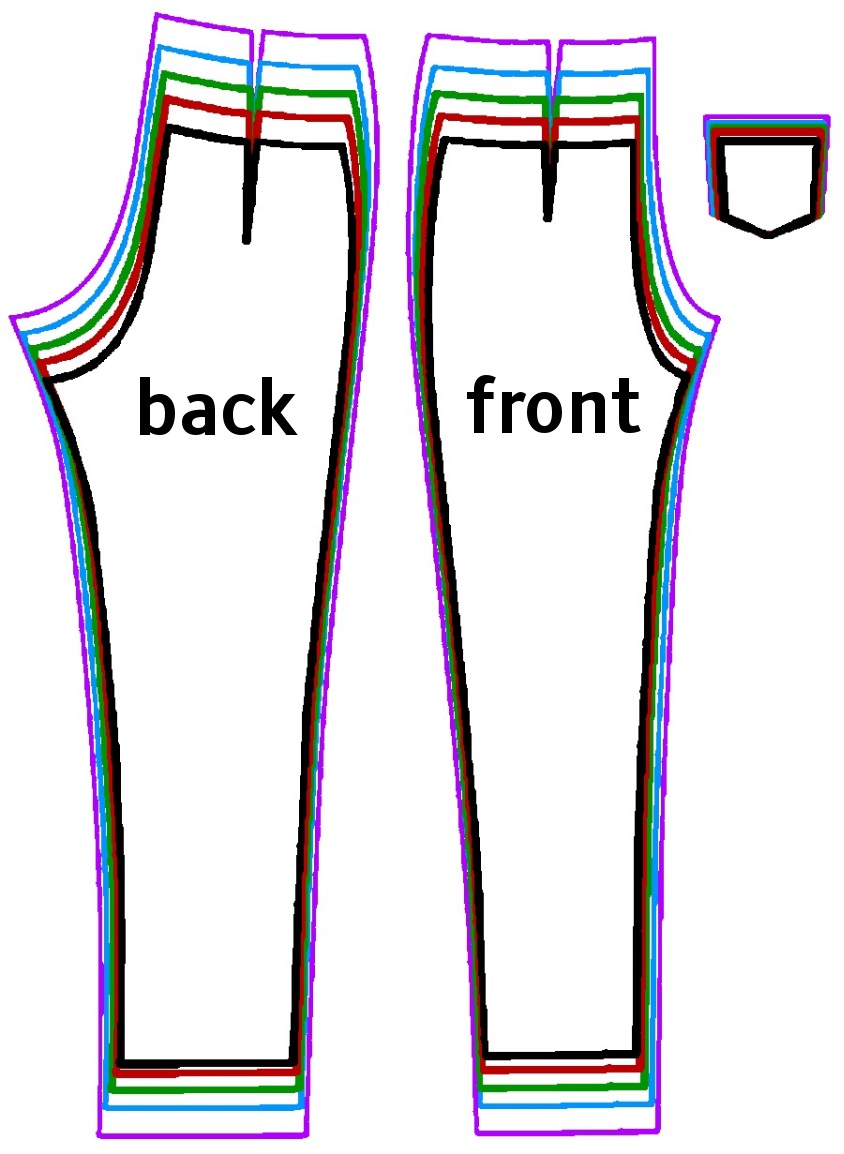}
          \label{classicPantsSewingPatternFig}
      }
\subfloat[]{
       \includegraphics[width =0.15\textwidth] {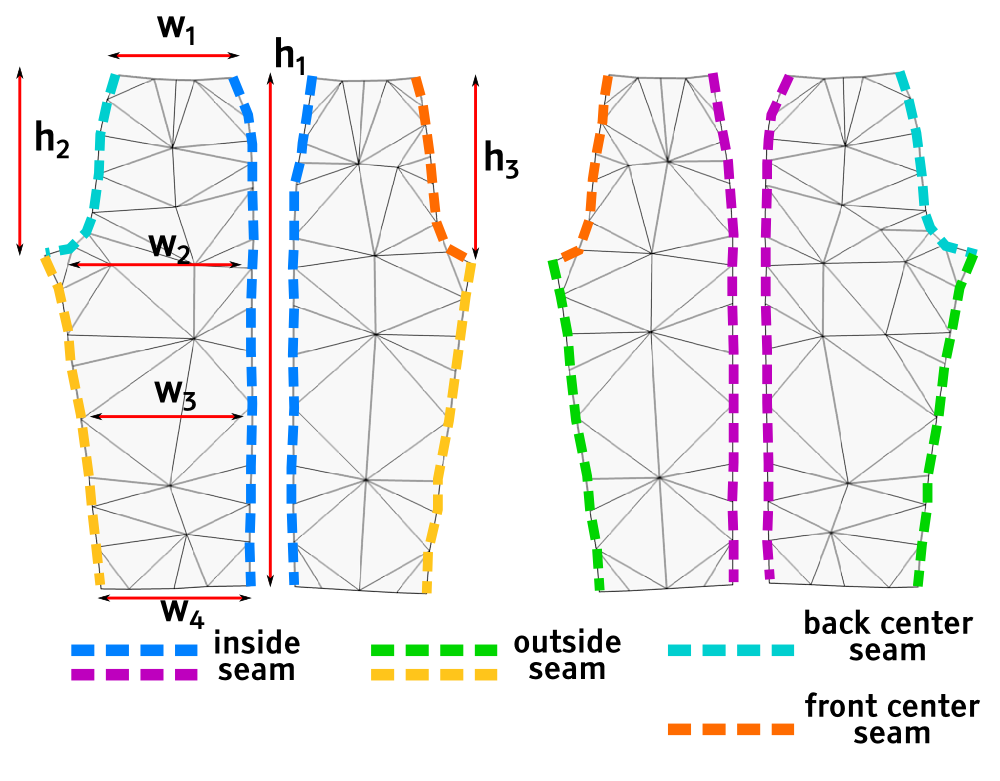}
          \label{pantsMSFig}
      }
\subfloat[]{
       \includegraphics[scale =0.09] {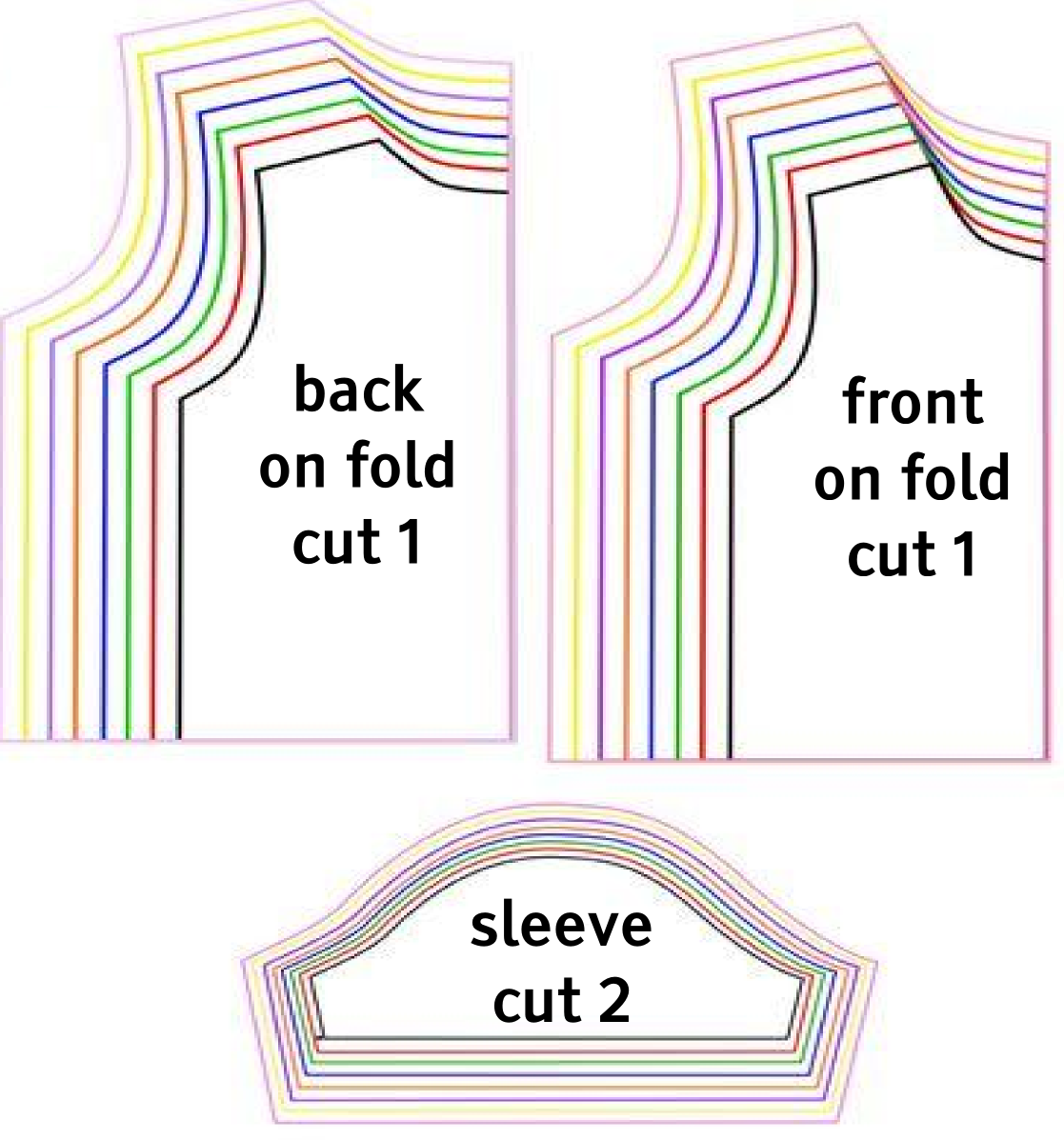}
          \label{classicTShirtSewingPatternFig}
      }
\subfloat[]{
       \includegraphics[scale=0.10] {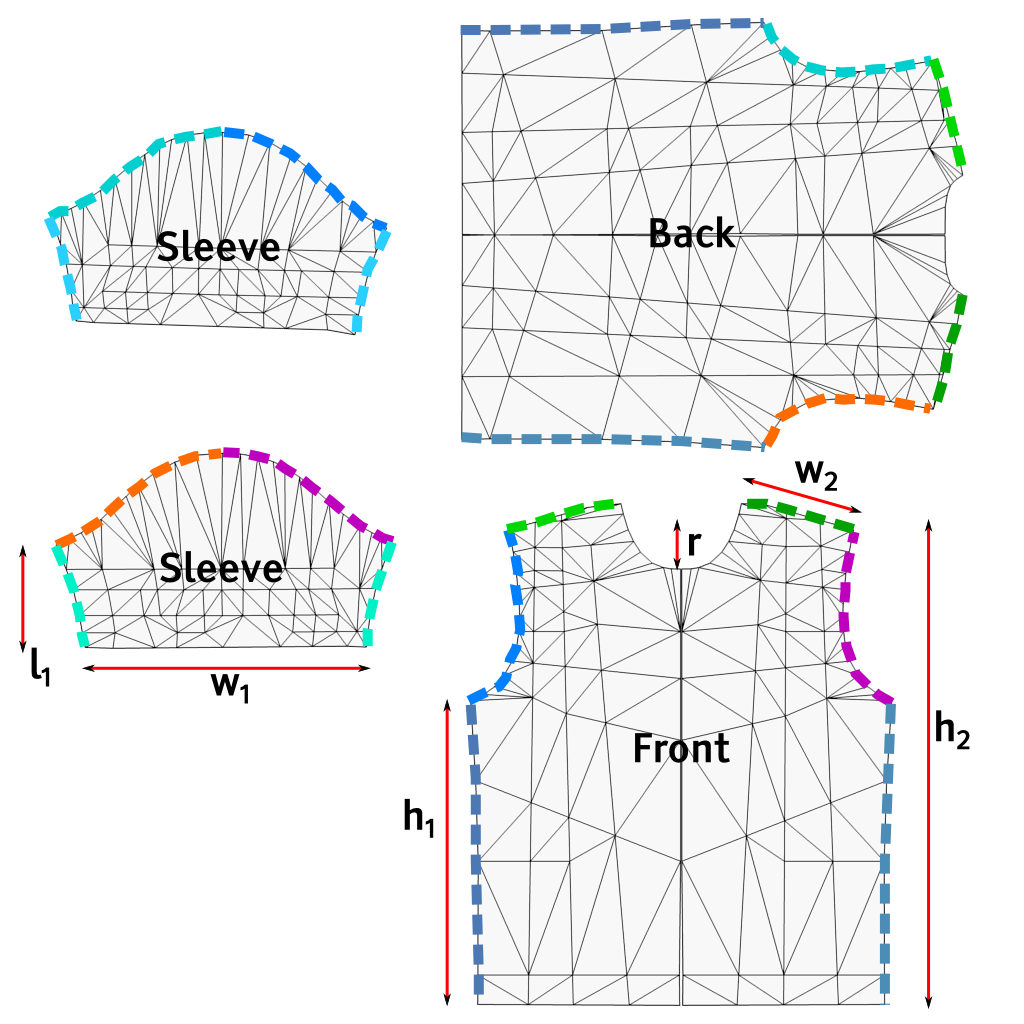}
          \label{tshirtMSFig}
      }
\end{center}
\vspace{-3mm}
\caption{{\bf Template sewing pattern and parameter space of a skirt, pants, and t-shirt. }(a) The classic circle sewing pattern of a skirt.
(b) Our parametric skirt template showing dashed lines for seams and the four parameters $<l_1,r_1,r_2,\alpha>$,
in which parameter $l_1$ is related to the waist girth and parameter $r_2$ is related to the length of the skirt.
(c) The classic pants sewing pattern.
(d) Our parametric pants template with seven parameters $<w_1,w_2,w_3,w_4,h_1,h_2,h_3>$.
(e) The classic t-shirt sewing pattern.
(f) Our parametric t-shirt template with six parameters $<r,w_1,w_2,h_1,h_2,l_1>$.
}
\label{sewingPaternFig}
\vspace{-3mm}
\end{figure*}

\vspace*{0.5em}
\noindent
{\bf Garment: }
For the clothing parameters, $\mathcal{C}$ is the set of material parameters including stretching stiffness and bending stiffness coefficients; $\E{U}$ is the 2D triangle mesh representing the garment's pattern pieces;
and $\E{V}$ is the 3D triangle mesh representation of the garment.
For each triangle of the 3D garment mesh $\E{V}$, there is a corresponding one 
in the 2D space $\E{U}$.     
For each mesh vertex $\E{x}\in\E{V}$, such as those  lying on a stitching seam in the garment, there might be multiple corresponding 2D vertices $\E{u}\in\E{U}$.
The parameter $\mathcal{G}$ is the set of parameters that defines the dimensions of the 2D pattern pieces. 
\begin{revision}
    We adopt the garment sizing parameters based on classic sewing patterns~\cite{barnfield2012the} shown in Fig.~\ref{classicSkirtSewingPatternFig},~\ref{classicPantsSewingPatternFig} and ~\ref{classicTShirtSewingPatternFig}, with the corresponding parameters defined in Fig.~\ref{skirtMSFig}, \ref{pantsMSFig}, 
and \ref{tshirtMSFig}, respectively.
\end{revision}
For example, we define the parameter $\mathcal{G}_{\text{pants}}={<w_1,w_2,w_3,w_4,h_1,h_2,h_3>}$ for pants, where the first four parameters define the waist, bottom, knee and ankle girth and the last three parameters indicate the total length, back upper, and front upper length.  
For each basic garment category, we can manually define this set of parameters $\mathcal{G}$. 
By manipulating the values of the parameters $\mathcal{G}$, garments of different styles and sizes can be modeled: capri pants vs. full-length pants, or tight-fitting vs. loose and flowy silhouettes. 
\begin{revision}
We use the material model developed by Wang et al.~\shortcite{wang2011data}. The material parameters $\mathcal{C}$ are the 18 bending and 24 stretching parameters.
\end{revision}

\vspace*{0.5em}
\noindent
{\bf Human Body: }
For the human body parameters, $\boldsymbol\theta$ is the set of joint angles that together parameterize the body pose, 
and $\E{z}$ is the set of semantic parameters that describe the body shape. 
We follow the PCA encoding of the human body shape presented in~\cite{hasler2009statistical}. 
The semantic parameters include gender, height, weight, muscle percentage, breast girth, waist girth, hip girth, thigh girth, calf girth, shoulder height, and leg length.  

\vspace*{0.5em}
Table~\ref{tab:notation} provides a list of formal definitions for the notation used in this paper.
\vspace*{0.5em}

\begin{table}[ht]
\vspace{-3mm}
\centering
\tbl{\bf Notation and definition of our method.}{
\resizebox{\columnwidth}{!}{
\begin{tabular}{lc}
    \hline
    NOTATION & DEFINITION \\
    \hline
    \hline
    $\mathcal{C}$ & material property parameters of the garment   \\
    % $\mathcal{C}'$ & material parameters after the garment parameter optimization \\
    $\mathcal{G}$ & garment (sizing) parameters   \\
    % $\mathcal{G}'$ & garment parameters after the garment parameter optimization \\
    $\mathcal{G}_{\text{pants}}$ & $<w_1,w_2,w_3,w_4,h_1,h_2,h_3>$ \\
    $\mathcal{G}_{\text{skirt}}$ & $<l_1,r_1,r_2,\alpha>$ \\   
    $\mathcal{G}_{\text{top}}$ & $<r,w_1,w_2,h_1,h_2,l_1>$ \\
    $\mathbf{U}$ & 2D triangle mesh representing garment pattern \\
    % $\mathbf{U}'$ & 2D triangle mesh after the garment parameter optimization \\
    $\mathbf{u}$ & vertex of the 2D garment pattern mesh \\
    $\mathbf{V}$ & 3D triangle surface mesh of the garment \\
    $\mathcal{V}$ & \begin{revision}simulated 3D triangle surface mesh of the garment \end{revision} \\
    $\mathbf{B}_{\text{body}}$ & skeleton of 3D human body mesh \\
    $\mathbf{B}_{\text{garment}}$ & skeleton of 3D garment mesh \\
    % $\hat{\mathbf{V}}$ & initially registered 3D triangle mesh of the garment \\
    $\mathbf{x}$ & vertex of the 3D triangle mesh \\
    % $\mathbf{x}_i^0$ & vertex of the template garment 3D triangle mesh \\
    $\mathcal{P}$ & \begin{revision} average wrinkle density of the 2D segmented garment in the image\end{revision} \\
    $\mathbf{k}$ & bending stiffness parameters \\
    $\mathbf{w}$ & stretching stiffness parameters \\
    $\mathbf{F}$ & deformation gradient of the deforming garment \\
    $\mathcal{K}$ & garment 3D mesh curvature measurement \\
    $\mathcal{S}$ & 2D garment silhouette \\ 
    $\Psi$ & bending energy of the garment \\
    $\Phi$ & stretching energy of the garment \\
    $\upsilon_{ij}$ & rigging weights of the 3D garment mesh \\
    $\boldsymbol\beta$ & joint angles of the skeleton of the 3D garment mesh \\
    \hline
    $\boldsymbol\theta$ & joint angles of the human body mesh \\
    $\mathbf{z}$ & semantic parameters of the human body mesh shape\\
    $\mathbf{p}$ & vertex of the 3D human body mesh \\
    $\mathbf{P}$ & 3D human body mesh \\
    % $\mathbf{p}^0_i$ & vertex of the mean 3D human body mesh \\
    $\mathbf{Z}_i$ & PCA basis of human body shape \\
    $\omega_i$ & rigging weights of the human body mesh \\
    \hline
    $\mathcal{D}_c$ & garment database \\
    $\mathcal{D}_h$ & human body database \\
    \hline
\end{tabular}}}
\label{tab:notation}
\end{table}

\section{Data Preparation}

This section describes the data preprocessing step.  We begin with the data 
representations for the garment and the human body, followed by a brief description of each preprocessing module.

\subsection{Data Representations}
The garment template database can be represented as a set $\mathcal{D}_c=\{<\mathcal{C}_i,\mathcal{G}_i,\E{U}_i,\E{V}_i, \E{B}_{\text{garment},i}>|i\in1,\cdots,N\}$, where $N$ is the number of garment templates. 
Each garment template consists of a 2D triangle mesh $\E{U}$ representing the sewing pattern, a 3D mesh $\E{V}$, a set of dimension parameters $\mathcal{G}$ for each pattern piece, the skeleton $\E{B}_{\text{garment}}$, and a set of material parameters $\mathcal{C}$.
The human body template database $\mathcal{D}_h=\{<\boldsymbol\theta_{j},\E{z}_{j},\E{B}_{\text{body},j}>|j\in1,\cdots,M\}$ consists of $M$ naked human body meshes with point to point correspondence.  
We use several human body datasets, including the SCAPE dataset~\cite{anguelov2005scape}, the SPRING dataset~\cite{yang2014semantic}, the TOSCA dataset~\cite{bronstein2008numerical,young2007calculus,bronstein2006efficient}, and the dataset from~\cite{hasler2009statistical}. 
\begin{revision}
Our garment template is defined in the same metric system as the human template
to scale the garments during the initial registration.
Each garment template and human template is rigged on a common skeleton with the same set of joints.

{\bf{Parameterized Garment Model:}}
Given the garment template database $\mathcal{D}_c$, each vertex $\mathbf{u}$ of the 2D garment pattern mesh is computed as
\begin{equation}
    \mathbf{u}(\mathcal{G}) = \sum_{i,g_i\in\mathcal{G}}^{|\mathcal{G}|}\nu_i(\mathbf{u}^0+g_i),
    \label{designPatternEqn}
\end{equation}
with $g_i$ as the $i^{\text{th}}$ 2D pattern sizing parameter in the set $\mathcal{G}$, $\nu_i$ is the weight associated with the vertex $\mathbf{u}$
and $\mathbf{u}^0$ is the vertex position of the 2D garment pattern template. 
\end{revision}
%To generate such a database, we use a dataset of single human dressed with $10$ different garments represented as $\{<\theta^a,\E{z}^a,\mathcal{C}_j^a,\E{x}_j^a>|j\in1,\cdots,10\}$ and another dataset of $100$ naked human bodies represented as $\{\PARAH{k}{b}|k\in1,\cdots,100\}$. We then generate a datum $<\theta_k^b,\E{z}_k^b,\mathcal{C}_j^a,\E{x}_j^a>$ for each $j,k$ pair. This is done by performing a quasistatic cloth simulation with the deforming human body (from $\PARAH{}{a}$ to $\PARAH{k}{b}$) as boundary condition.

{\bf{Parameterized Human Model:}}
Given the body database $\mathcal{D}_h$, we extract a statistical shape model for human bodies. Under this model, each world space vertex position $\E{p}$ on the human body is parameterized as
\begin{eqnarray}
    \E{p}(\boldsymbol\theta,\E{z})=\sum_i^{|\mathbf{B}_{\text{body}}|}\omega_i\mathcal{T}_{\mathcal{B}_i}(\boldsymbol\theta)\left(\E{p}^0+\E{Z}_i\E{z}\right),
\end{eqnarray}
which is a composition of a linear blend skinning model \cite{kavan2010fast} and an active shape model \cite{zhao2003face}. Here $\omega_i$ and $\mathcal{B}_i$ are the set of weights and bones associated with the vertex $\mathbf{p}$.  $\mathcal{T}_{\mathcal{B}_i}$ is the transformation matrix of bone $\mathcal{B}_i$.  $\E{p}^0$ and $\E{Z}_i$ are the mean shape and active shape basis at the rest pose, respectively. The basis $\E{Z}_i$ is calculated by running PCA \cite{hasler2009statistical} on $\mathcal{D}_h$. 

\subsection{Preprocessing}
\label{preprocessingSec}
Our preprocessing step consists of: a) human body reconstruction to recover the human body shape and pose from the input image, b) garment parsing to estimate the locations and types of garments depicted in the image, and c) parameter estimation to compute the sizing and fine features of the parsed garments.

{\bf Human Body Reconstruction:}
Our human body recovery relies on limited user input. 
The user helps us identify the 14 human body joints and the human body silhouette.
With the identified joints, a human body skeleton is recovered using the method presented in~\cite{taylor2000reconstruction}:
the semantic parameters $\mathbf{z}$ are optimized to match the silhouette. 
In this step, we ignore the camera scaling factor. 

{\bf{Garment Parsing:}}  
We provide two options for garment parsing.
The first uses the automatic computer vision technique presented in~\cite{yamaguchi2013paper}.
This approach combines global pretrained parse models with local models learned from nearest neighbors and transferred parse masks to estimate the types of garments and their locations on the person.
The second option requires assistance from the user. 
Given the image $\Omega$, we extract the clothing regions $\Omega_{b,h,g}$ by performing a two-stage image segmentation guided by user sketch. In the first stage, a coarse region boundary is extracted using a graph cut algorithm~\cite{li2004lazy}. Then, the region is refined via re-clustering~\cite{levin2008closed}.
%\begin{figure}[hbt]
%\centering
%\includegraphics[width=0.48\textwidth]{./figures/matting.eps}
%\caption{The input image (left) and the extracted human body part (white) and cloth region (red) guided by user sketch.}
%\label{fig:cluster}
%\vspace{-3mm}
%\end{figure}

{\bf{Image Information Extraction:}}
\begin{revision}
Given the segmentation of the garment $\Omega_g$, the next step is to convert it to pixel-level garment silhouette $\mathcal{S}$ and compute the average wrinkle density $\mathcal{P}$. 
Instead of using the wrinkle density for each part of the garment, the average wrinkle density encodes the overall material properties of it. 
We extract the average wrinkle density $\mathcal{P}$ from the garment images using an improved implementation of ~\cite{popa2009wrinkling}. 
We first detect edges using Holistically-Nested edge detection~\cite{xie2015holistically} and then smooth the edges by fitting them to low-curvature quadrics. 
We smooth edges split during the detection phase by merging those that have nearby endpoints and similar orientations. 
Finally, we form 2D folds by matching parallel edges. 
%This pairing favors long edges with bright regions in between. 
Edges not part of a pair are unlikely to contribute to a fold and are discarded. 
The average number of wrinkles per area is the average wrinkle density $\mathcal{P}$.\end{revision}

\subsection{Initial Garment Registration}
\label{initialGarmentRegistrationSec}
Our initial clothing registration step aims to dress our template garment onto a human body mesh of any pose or shape.   
We optimize the vertex positions of the 3D mesh, $\E{x}$, of the template clothing based on the human body mesh parameters $<\boldsymbol\theta,\E{z}>$.
In this step, we ignore the fit of the clothing on the human body (this step is intended to fix the 2D triangle mesh $\E{U}$). 
We follow the method proposed in~\cite{brouet2012design} for registering a template garment to a human body mesh with a different shape.  
However, their method is unable to fit the clothing to human meshes with varying poses; we extend their approach by adding two additional steps. 

\begin{figure}[!ht]
\centering
\hspace{-3mm}
\subfloat[]{
       \includegraphics[width =0.12\textwidth] {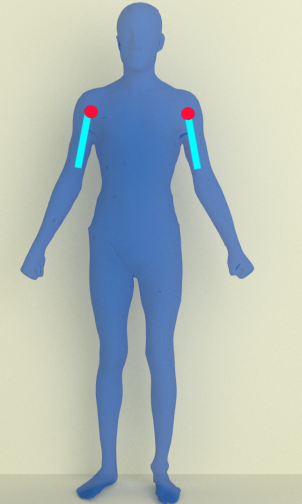}
          \label{humanSkelFig}
      }\hspace{-3mm}
 \subfloat[]{
        \includegraphics[width =0.12\textwidth] {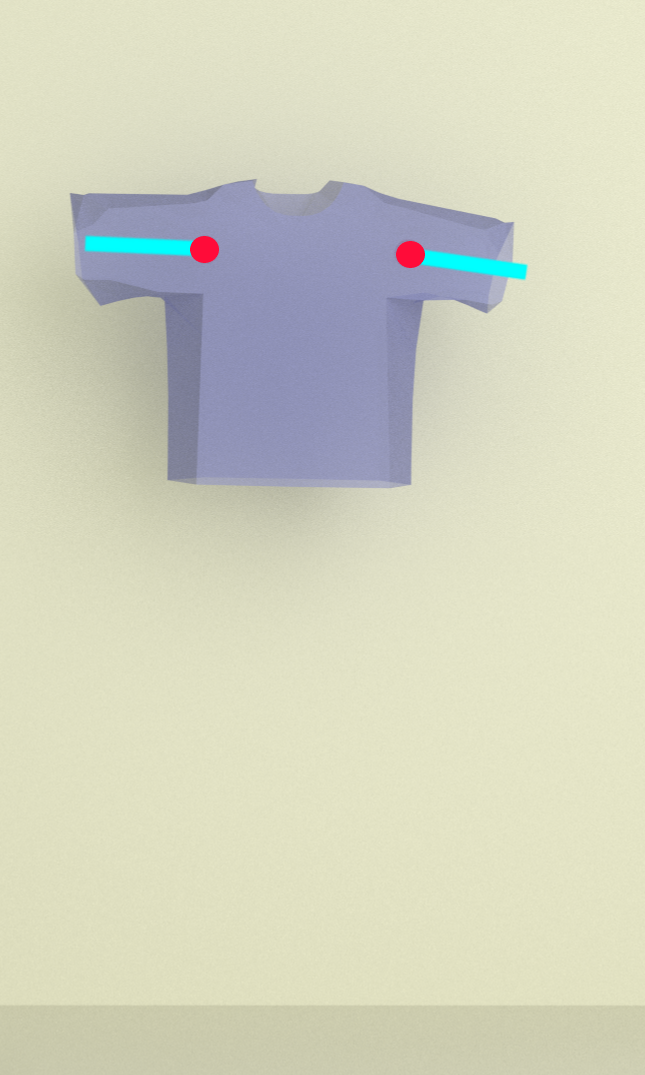}
          \label{clothingSkelFig}
    }\hspace{-3mm}
\subfloat[]{
       \includegraphics[width =0.12\textwidth] {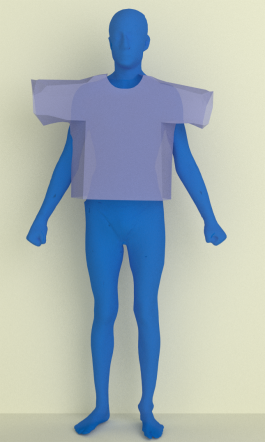}
          \label{humanClothingFig}
      }\hspace{-3mm}
 \subfloat[]{
        \includegraphics[width =0.12\textwidth] {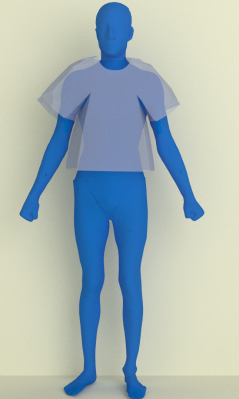}
          \label{humanClothingRigFig}
    }\hspace{-3mm}
\caption{{\bf Initial garment registration process. }(a) The human body mesh with the skeleton joints shown as the red sphere and the skeleton of the arm shown as the cyan cylinder.
(b) The initial t-shirt with the skeleton joint shown as the red sphere and the skeleton of the sleeve part of it shown as the cyan cylinder. 
(c) The t-shirt and the human body mesh are aligned by matching the joints.
(d) The result after aligning the skeleton and removing the interpenetrations.} 
\label{eigenFuncVisualFig}
\end{figure}

The first step requires the alignment of the joints $\mathcal{Q}_c$ of the template garment skeleton with the joints $\mathcal{Q}_h$ of the human body mesh skeleton, 
as shown in Fig.~\ref{eigenFuncVisualFig}.  
Each joint $\mathbf{q}\in\mathcal{Q}_c$ of the garment has one corresponding joint $\mathbf{t}\in\mathcal{Q}_h$ of the human body mesh.
We denote the number of joints of the garment as $K_c$.
This step is done by applying a rigid body transformation matrix $\mathbf{T}$ on the joint of the garment, where $\mathbf{T}$ minimizes the objective function 
\begin{equation}
% This is incorrect -- ML fixed!!!
%    E_{\text{joints}}=\argmin_{\mathbf{T}}\sum_{i=0,\mathbf{q}_i\in\mathcal{Q}_c,\mathbf{t}_i\in\mathcal{Q}_h}^{K_c}\|\mathbf{T}\mathbf{q}_i-\mathbf{t}_i\|
%
    \sum_{i=0,\mathbf{q}_i\in\mathcal{Q}_c,\mathbf{t}_i\in\mathcal{Q}_h}^{K_c}\|\mathbf{T}\mathbf{q}_i-\mathbf{t}_i\|^2
\end{equation}
Next, we need to fit this transformed 3D template garment mesh onto the human body mesh with pose described by parameter $\boldsymbol\theta$, the vector of the angles of the joints. 
Our template garment is then deformed according to $\boldsymbol\theta$.  
We denote the vector $\boldsymbol\beta$ as the joint angles of the template garment mesh. 
We set the value of the vector $\beta_i$ to the value of the corresponding joint angle $\theta_i$ of the human body mesh. 
Then we compute the 3D garment template mesh such that it matches the pose of the underlying human body mesh according to this set of joint angles $\boldsymbol\beta$ by,
\begin{equation}
    \mathbf{x}_i(\boldsymbol\beta)=\sum_j\upsilon_{ij}\mathcal{T}_{\mathcal{B}_j}(\boldsymbol\beta)\E{x}^0,
\end{equation}
where $\upsilon_{ij}$ is the weight of bone $\mathcal{B}_j$ on vertex $\mathbf{x}_i$ and $\mathcal{T}_{\mathcal{B}_j}$ is the transformation matrix of the bone $\mathcal{B}_j$. An example result is shown in Fig.~\ref{humanClothingFig}.

\begin{figure}[!hbt]
\centering
\includegraphics[width=0.48\textwidth]{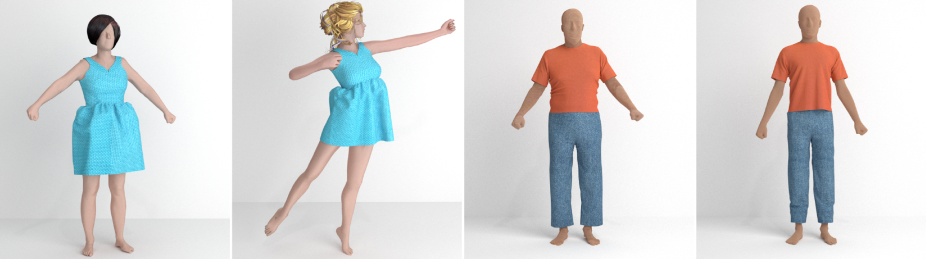}
\caption{{\bf Initial garment registration results.} We fit garments to human bodies with different body shapes and poses.}
\label{initialRegistrationResultFig}
\end{figure}

The final step is to remove collisions between the garment surface mesh and the human body mesh. 
This step is similar to the ICP algorithm proposed by Li et al.~\shortcite{li2008global}. 
We introduce two constraints: rigidity and non-interception. 
The deformation of the clothing should be as-rigid-as-possible~\cite{igarashi2005rigid}.  
After this step, we have an initial registered garment with a 3D mesh $\hat{\mathbf{V}}(\mathbf{T},\boldsymbol\theta)$ that matches the underlying human pose and is free of interpenetrations with the human body. 
We show our initial garment registration results in Fig.~\ref{initialRegistrationResultFig}.

\section{Image-Guided Parameter Identification}
\label{garmentParameterSec}

In this section, we explain the step-by-step process of extracting garment 
material and sizing parameters ${<\mathcal{C},\mathcal{G}>}$ from an image.

\subsection{Overview}

Starting from our 2D triangle mesh $\mathbf{U}$ of the pattern pieces, we select garment parameters based on the sizing and detailed information ${<\mathcal{S},\mathcal{P}>}$ estimated from the source image. 
\begin{revision}
In this step, we adjust the garment material and sizing parameters $<\mathcal{C},\mathcal{G}>$ but fix the 3D mesh $\hat{\mathbf{V}}(\mathbf{T},\boldsymbol\theta)$ (computed from Sec.~\ref{initialGarmentRegistrationSec}) to obtain the garment that best matches the one shown in the image. 
We need two specific pieces of information from the image: the pixel-level garment silhouette  $\mathcal{S}$ and the average wrinkle density $\mathcal{P}$ of the clothing. 
For example, for a skirt, we need to estimate the waist girth and the length of the skirt from the image. 
\end{revision}
Using this information, we initialize the garment sizing parameters $\mathcal{G}$. 
%The wrinkle density, derived from the wrinkle density $\mathcal{P}$ of the skirt provides important hints about material properties, such as stretching and bending. 
Based on the wrinkle information computed from the image, we optimize both the fabric material parameters $\mathcal{C}$ and the sizing parameters of the garment pattern $\mathcal{G}$.  
%Through dynamic cloth simulation we get the final clothings that match the one in the single-view image. 

\subsection{Garment Types, Patterns, and Parameters}

For basic garment types, such as skirts, pants, t-shirts, and tank tops, we use one template pattern for each. 
We modify the classic sewing pattern according to the parameters $\mathcal{G}$. 
By adjusting the garment parameters $\mathcal{G}$ and fabric material parameters $\mathcal{C}$, we recover basic garments of different styles and sizes. 
The classic circle skirt sewing pattern is shown in Fig.~\ref{classicSkirtSewingPatternFig}. 
Our parametric space, which is morphed from this circle sewing pattern, is shown in Fig.~\ref{skirtMSFig}. 
For the skirt pattern, there are four parameters to optimize: $\mathcal{G}_{\text{skirt}}=<l_1,r_1,r_2,\alpha>$. 
The ratio between the parameter $l_1$ and $r_2$ is constrained by the waist girth and skirt length information extracted from the image. 
The other two parameters, $r_1$ and $\alpha$, are constrained by the wrinkle density.  With different garment parameters, skirts can vary from long to short, stiff to soft, and can incorporate more or fewer pleats, enabling us to model a wide variety of skirts from a single design template.

Similarly for pants, the classic sewing pattern and our template pattern pieces are shown in Fig.~\ref{classicPantsSewingPatternFig} and Fig.~\ref{pantsMSFig}.
There are seven parameters for the dimensions of the pants template: ${\mathcal{G}_{\text{pants}}=<w_1,w_2,w_3,w_4,h_1,h_2,h_3>}$ with the first four parameters describing the waist, bottom, knee, and ankle girth,
and the last three parameters representing the total, back-upper and front-upper lengths.
The t-shirt sewing pattern is shown in Fig.~\ref{classicTShirtSewingPatternFig}, and our parametric t-shirt pattern is shown in Fig.~\ref{tshirtMSFig} with the garment parameters $\mathcal{G}_{\text{tshirt}}=<r,w_1,w_2,h_1,h_2,l_1>$. 
Among the parameters $\mathcal{G}_{\text{tshirt}}$, parameter $r$ describes the neckline radius, $w_1$ describes the sleeve width, $w_2$ describes the shoulder width, $h_1$ describes the bottom length, $h_2$ describes the total length, and $l_1$ describes the length of the sleeve.
\begin{figure}[!bht]
\vspace{-6mm}
\hspace{-8mm}
\begin{center}
\subfloat[]{
       \includegraphics[scale=0.13] {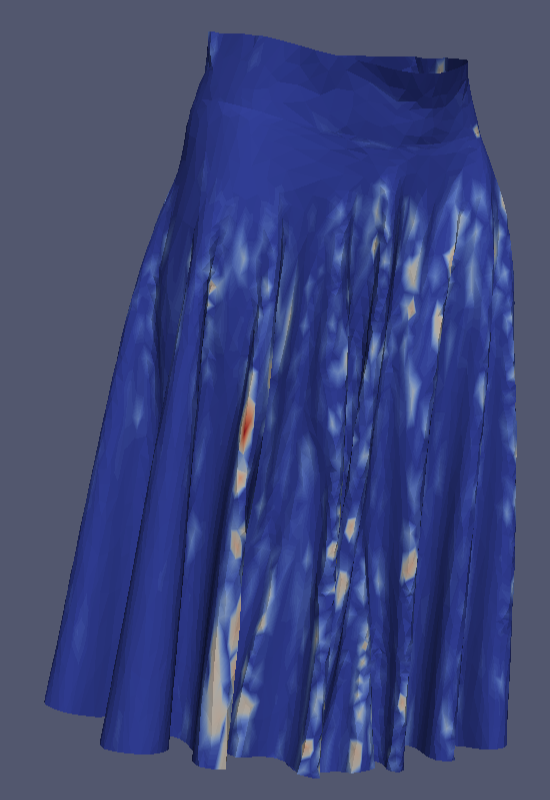}
          \label{beforeOptSkirtFig}
      }
\subfloat[]{
       \includegraphics[scale=0.13] {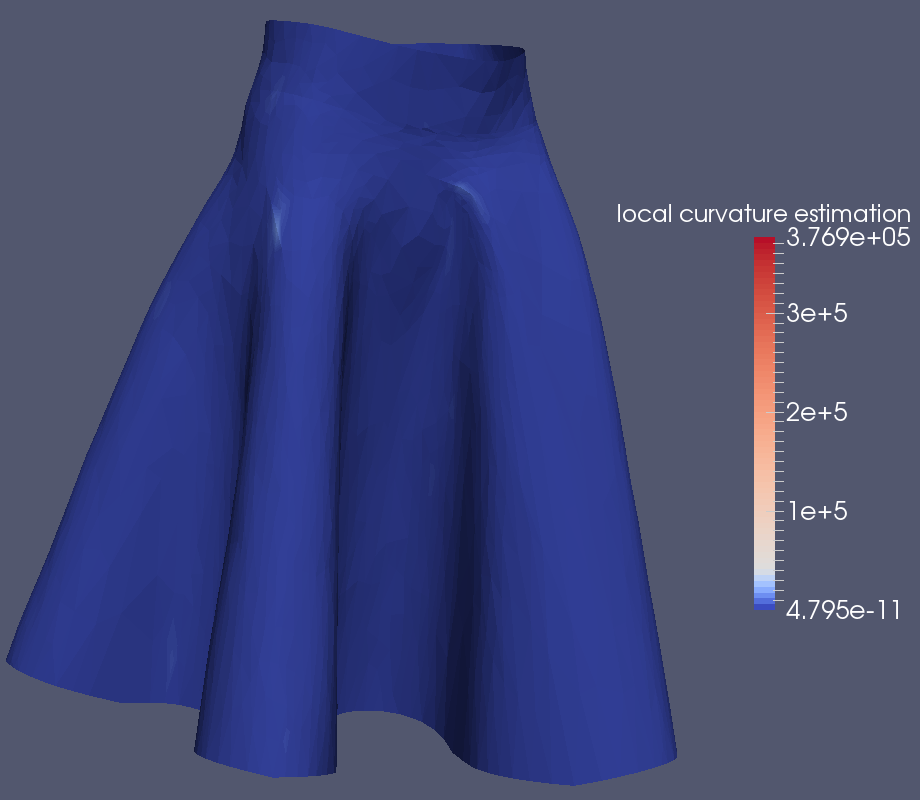}
          \label{afterOptSkirtFig}
      }
\subfloat[]{
       \includegraphics[scale=0.17] {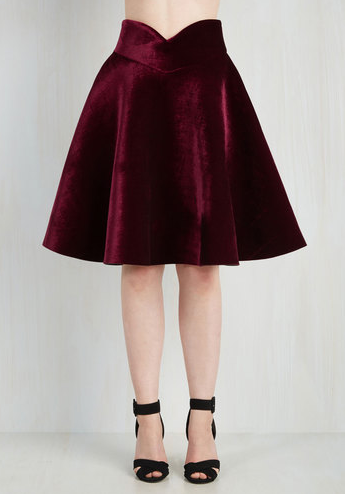}
          \label{targetSkirtFig}
      }
\end{center}
\vspace{-3mm}
\caption{{\bf Material parameter identification results. }(a) The local curvature estimation before optimizing the bending stiffness coefficients (using the cool-to-hot color map). 
    (b) The local curvature estimation after the optimization. 
    (c) The original garment image {\em \protect\cite{skirt}~$\copyright$}.
} 
\label{optResultsFig}
\end{figure}

%Many wrinkle patterns of garments, especially in skirts and dresses, are formed due to the sewing patterns. 
Different sewing patterns result in very different garments.
Traditional sewing techniques form skirt wrinkles by cutting the circular portion of the pattern. 
To simulate this process but make it generally applicable, we modify the parameter $\mathcal{G}$, which achieves the same effect. 
In addition to the differences created by sewing patterns, professional garment designers also take advantage of cloth material properties to produce different styles of clothing. 
We tune the bending stiffness coefficients and stretching stiffness coefficients in $\mathcal{C}$ to simulate this cloth selection process.  

\begin{revision}
\subsection{From Wrinkle Density to Material Property}

One of the key insights in this work is the identification of fabric materials based on wrinkles and
folds, because different fabric stiffness produce varying wrinkle/folding patterns.  
We characterize the wrinkles and folds using their local curvatures.
The first step is to map the wrinkle density $\mathcal{P}$ (computed in Sec.~\ref{preprocessingSec}) to the average local curvature $\mathcal{K}$. 
\end{revision}

We recover the garment material parameter $\mathcal{C}'$ by minimizing the average local curvature differences between our recovered garment $\mathcal{K}(\mathcal{C},\mathcal{G})$ and the reference garment $\mathcal{K}(\mathcal{P})_{\text{target}}$
\begin{revision}
\begin{equation}
    \mathcal{C}' = \argmin_{\mathcal{C}}\|\mathcal{K}(\mathcal{C},\mathcal{G})-\mathcal{K}(\mathcal{P})_{\text{target}}\|^2.
\end{equation}
The reference garment average local curvature $\mathcal{K}(\mathcal{P})_{\text{target}}$ is computed by linear interpolation.
%The projection process includes two steps: estimation and linear interpolation. 
We first approximate the average local curvature threshold for the sharp wrinkles and smooth folds.
The average local curvature threshold for one sharp wrinkle is up to $10^5$ $\text{m}^{-1}$ and that for smooth folds is close to $10^{-4}$ $\text{m}^{-1}$. 
Sharp wrinkles or large folds are determined by the density of the extracted 2D wrinkles. 
The density of the extracted 2D wrinkles ranges from $1$ $\text{m}^{-2}$ to 
$50$ $\text{m}^{-2}$ based on our observation.
The interpolation process (with the linear interpolation function $\mathcal{I}$) is
\begin{equation}
	\mathcal{K}(\mathcal{P})_{\text{target}}=\mathcal{I}(\mathcal{P}),
\end{equation}
with the linear interpolation function $\mathcal{I}(1) = 10^{-4}$ and $\mathcal{I}(50) = 10^5$.
Local curvature estimation of $\kappa$ at each vertex is computed based on the bending between the two neighboring triangles sharing the same edge.  
For each vertex $\mathbf{x}$ of the two triangles that share an edge $\mathbf{e}$, the local curvature $\kappa$ is computed following the approach from Wang et al.~\shortcite{wang2011data} and Bridson et al.~\shortcite{bridson2003simulation}
\begin{equation}
    \kappa=\|\sin(\alpha/2)(h_1+h_2)^{-1}|\mathbf{e}|\mathbf{x}\|,
\end{equation}
where $h_1$ and $h_2$ are the heights of the two triangles that share the edge $\mathbf{e}$ and $\alpha$ is the supplementary angle to the dihedral angle between the two triangles.
The corresponding bending force $\mathbf{f}_{\text{bend}}$ for each vertex $\mathbf{x}$ is computed as 
\begin{equation}
    \mathbf{f}_{\text{bend}}=k\sin(\alpha/2)(h_1+h_2)^{-1}|\mathbf{e}|\mathbf{x},
\end{equation}
where $k$ is the bending stiffness coefficient.  

Stretching also affects the formation of wrinkles. 
Each triangle $<\mathbf{u}_0,\mathbf{u}_1,\mathbf{u}_2>$ in the 2D template mesh is represented as $D_m=\begin{bmatrix}\mathbf{u}_1-\mathbf{u}_0\\\mathbf{u}_2-\mathbf{u}_0\end{bmatrix}$, and each triangle in the 3D garment mesh $<\mathbf{x}_0,\mathbf{x}_1,\mathbf{x}_2>$ is represented as $d_m=\begin{bmatrix}\mathbf{x}_1-\mathbf{x}_0\\\mathbf{x}_2-\mathbf{x}_0\end{bmatrix}$. 
    The stretching forces $\mathbf{f}_{\text{stretch}}$ are computed by differentiating the stretching energy $\Psi$, which depends on the stretching stiffness parameter $\mathbf{w}$, the deformation gradient $\mathbf{F}=d_mD_m^{-1}$, and the Green strain tensor $\mathbf{G}=\frac{1}{2}(\mathbf{F}^T\mathbf{F}-\mathbf{I})$ against the vertex 3D position $\mathbf{x}$
\begin{equation}
    \mathbf{f}_{\text{stretch}}=\frac{\partial\Psi(\mathbf{w},\mathbf{F})}{\partial\mathbf{x}}.
\end{equation}
\vspace{-4mm}

The sizing and style of the garment described by the parameter $\mathcal{G}$ obtained from the parsed garment are matched by minimizing the silhouette which is a 2D polygon differences between our recovered garment $\mathcal{S}(\mathcal{C},\mathcal{G})$ and the reference garment silhouette $\mathcal{S}_{\text{target}}$
\begin{equation}
    \mathcal{G}'=\argmin_{\mathcal{G}}\|\mathcal{S}(\mathcal{C},\mathcal{G})-\mathcal{S}_{\text{target}}\|^2.
\end{equation}
The distance between two polygons is computed by summing up the distances between each point in polygon $\mathcal{S}(\mathcal{C},\mathcal{G})$ to the other polygon $\mathcal{S}_{\text{target}}$.
To compute the 2D silhouette $\mathcal{S}(\mathcal{C},\mathcal{G})$, we first project the simulated garment 3D mesh ${\mathcal{V}}(\mathcal{C},\mathcal{G}, \hat{\mathbf{V}}(\mathbf{T},\boldsymbol\theta))$ onto the the 2D image with the projection matrix $\mathbf{H}$.
Then compute the 2D polygon enclosing the projected points.
%The projection matrix $\mathbf{H}$ maps $\mathbf{V}(\mathcal{C},\mathcal{G})$ on to the 2D image $\Omega$. 
The process is expressed as 
\begin{equation}
\mathcal{S}(\mathcal{C},\mathcal{G})=f(\mathbf{H}\mathcal{V}(\mathcal{C},\mathcal{G},\hat{\mathbf{V}}(\mathbf{T},\boldsymbol\theta))),
    \label{silhouetteComputeEqn}
\end{equation}
with $f$ as the method that convert the projected points to a 2D polygon.
We ignore the camera scaling factor in this step since the input is a single-view image. 
It is natural to scale the recovered clothing and human body shape as a postprocessing step. 

Combining these two objectives, the objective (energy) function is expressed as 
\begin{equation}
    \begin{aligned}
        \mathbf{E} = & \|\mathcal{K}(\mathcal{C},\mathcal{G})-\mathcal{K}(\mathcal{P})_{\text{target}}\|^2+\\
                    & \|\mathcal{S}(\mathcal{C},\mathcal{G})-\mathcal{S}_{\text{target}}\|^2.
    \end{aligned}
\label{objectiveFuncEqn}
\end{equation}
\end{revision}
\subsection{Optimization-based Parameter Estimation}

The optimization is an iterative process (given in Algorithm~\ref{paramAlg}), alternating 
between updates for the garment sizing and material parameters, $\mathcal{G}$ and $\mathcal{C}$.
We found that the value of the objective function is more sensitive to the cloth material properties $\mathcal{C}$ than to the garment parameter $\mathcal{G}$, so we maximize the iterations when optimizing for $\mathcal{C}$, fixing $\mathcal{G}$. 
The optimization of parameter $\mathcal{C}$ is coupled with the cloth dynamics. 
The underlying cloth simulator is based on the method proposed in~\cite{narain2012adaptive}.
We drape the initial fitted garment onto the human body mesh. 
The garment is in the dynamic state and subject to gravity. 
%The physically based draping operation is an optimization process. 
%When the bending energy $\Phi(\mathbf{k},\kappa)$ and the stretching energy $\Psi(\mathbf{w},\mathbf{F})$ are minimized, the clothing is in the rest state.  
We couple our parameter estimation with this physically based simulation process. 
Before the simulation, we change the cloth material parameters $\mathcal{C}$ so that when in static state the average of the local curvature $\kappa$ matches the targeting threshold $\mathcal{K}_{\text{target}}$.
That is to say, 
%when the simulator minimizes the energy $\Phi$, 
our optimizer minimizes $\|\mathcal{K}-\mathcal{K}_{\text{target}}\|^2$ by changing the bending stiffness parameters $\mathbf{k}$ and stretching stiffness parameters $\mathbf{w}$.

We apply the L-BFGS~\cite{liu1989limited} method for our material parameter optimization. 
When the clothing reaches a static state, the optimizer switches to optimizing parameter $\mathcal{G}$. 
The optimizer for the parameter $\mathcal{G}$ is not coupled with the garment simulation. 
The objective function is evaluated when the clothing reaches the static state. 
We adopt the Particle Swarm Optimization (PSO) method~\cite{kennedy2010particle} for the parameter $\mathcal{G}$ optimization. 
\begin{revision} The PSO method is known to be able to recover from local minima, making it the ideal method for some of the non-convex optimization problems.  \end{revision}
We use $40$ particles for the parameter estimation process. 
The alternating process usually converges after four steps.  
One example result of the garment parameter process is shown in Fig.~\ref{optResultsFig}.

%Before we start the optimization of Eqn.~\ref{objectiveFuncEqn}, 
We constrain the cloth material parameter space.   
We use the ``Gray Interlock'' 
presented in~\cite{wang2011data}, which is composed of $60\%$ cotton and $40\%$ polyester,
as the ``softest'' material, meaning it bends the easiest. 
We multiply the bending parameters of this material by $10^2$ to give the ``stiffest'' material based on our experiments.  
Our solution space is constrained by these two materials, and we initialize our 
optimization with the ``softest'' material parameters. 

\begin{revision}
\begin{algorithm}
\caption{Garment Parameter Identification}
\label{garmentRecoveryAlgorithm}
    \begin{algorithmic}[1]
        \Procedure{SizingParamIdentification}{$\mathcal{G}$, $\mathcal{S}_{\text{target}}$}
        \State Compute silhouette $\mathcal{S}(\mathcal{C},\mathcal{G})$ using Eqn.~\ref{silhouetteComputeEqn}
        \State Minimize $\|\mathcal{S}(\mathcal{C},\mathcal{G})-\mathcal{S}_{\text{target}}\|^2$ using PSO 
        \State $\mathcal{G}'=\argmin_{\mathcal{G}}\|\mathcal{S}(\mathcal{C},\mathcal{G})-\mathcal{S}_{\text{target}}\|^2$
        \State Update the 2D mesh $\mathbf{U}'$ using Eqn.~\ref{designPatternEqn}
        \State \Return $\mathcal{G}'$, $\mathbf{U}'$
        \EndProcedure
        \Procedure{MaterialParamIdentification}{$\mathcal{C}$, $\mathcal{K}_{\text{target}}$}
        \State Compute local curvature $\mathcal{K}(\mathcal{C})$
        \State Minimize $\|\mathcal{K}(\mathcal{C})-\mathcal{K}(\mathcal{P})_{\text{target}}\|^2$ using L-BFGS Method 

        \State $\mathcal{C}' = \argmin_{\mathcal{C}}\|\mathcal{K}(\mathcal{C},\mathcal{G})-\mathcal{K}(\mathcal{P})_{\text{target}}\|^2$
        \State \Return $\mathcal{C}'$
        \EndProcedure
        \Procedure{Main}{ $\mathcal{C}$, $\mathcal{G}$, $\mathbf{U}$, $\epsilon$}
        \While{$\mathbf{E}>\epsilon$} \hspace*{0.5in} // $\mathbf{E}$ as defined in Eqn.~\ref{objectiveFuncEqn}
        \State \Call{MaterialParamIdentification}{$\mathcal{C}$}
        \State \Call{SizingParamIdentification}{$\mathcal{G}$}
        \EndWhile
        \State \Return $\mathcal{G}'$,$\mathcal{C}'$,$\mathbf{U}'$
        \EndProcedure
    \end{algorithmic}
\label{paramAlg}
\end{algorithm}
\end{revision}

\section{Joint Material-Pose Optimization}
\label{sec:fitting}

\subsection{Optimal Parameter Selection}
\label{jointOptimizationSec}
The parameter identification step provides us with the initial recovered garment described by the set of material and sizing parameters $<\mathcal{C}',\mathcal{G}'>$. 
\begin{revision}
Many realistic garment wrinkles and folds, however, are formed due to the underlying pose of the human body, especially wrinkles that are located around human joints. 
Therefore, in this step, we further refine our results by optimizing both the pose parameters of the human body $\boldsymbol\theta$ and the material properties of the cloth $\mathcal{C}'$. 
The optimization objective for this step is
\end{revision}
\begin{equation}
    \mathbf{E}_{\text{joint}}=\|\mathcal{K}(\mathcal{C}',\boldsymbol\theta)-\mathcal{K}(\mathcal{P})_{\text{target}}\|^2.
    \label{jointObjectiveFuncEqn}
\end{equation}

\begin{figure}[h]
\centering
\hspace{-3mm}
     \includegraphics[width=0.5\textwidth]{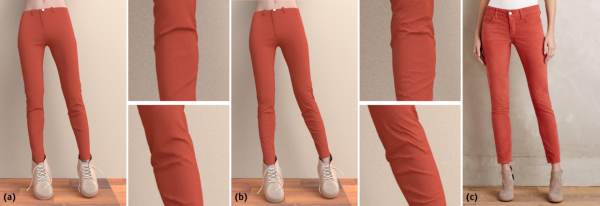}
\caption{{\bf Joint Material-Pose Optimization results.} (a) The pants recovered prior to the joint optimization. 
(b) The recovered pants after optimizing both the pose and the material properties. 
The wrinkles in the knee area better match with those in the original image. 
(c) The original pants image {\em \protect\cite{pants}~$\copyright$.}}
\label{jointFittingFig}
\vspace{-3mm}
\end{figure}

The optimization process (shown in Algorithm~\ref{jointAlg}) is similar to the garment parameter identification step, alternating between updating the pose parameter $\boldsymbol\theta$ and the material parameters $\mathcal{C}'$. 
\begin{revision}
    We use Particle Swarm Optimization method~\cite{kennedy2010particle}.
% minimize the objective as described in Eqn.~\ref{jointObjectiveFuncEqn}.  
\end{revision}
The objective function (Eqn.~\ref{jointObjectiveFuncEqn}) is more sensitive to the pose parameter $\boldsymbol\theta$ than to the material parameters $\mathcal{C}'$. 
%Similarly, we maximize the iteration of the optimization of the pose parameter $\boldsymbol\theta$. 
We constrain the optimization space of parameter $\boldsymbol\theta$ by confining the rotation axis to only the three primal axes. 
An example of our joint material-pose optimization method is shown in Fig.~\ref{jointFittingFig}.
%Combined previous steps, the entire approach is shown in Algorithm.~\ref{garmentRecoveryAlgorithm}.

\begin{revision}
\begin{algorithm}
\caption{Joint Pose-Material Parameter Identification}
    \begin{algorithmic}[1]
    \Procedure{Main}{ $\mathcal{C}'$, $\mathcal{G}'$, $\boldsymbol{\theta}$, $\epsilon$}
    \While{$\mathbf{E}_{\text{joint}}>\epsilon$} \hspace*{0.5in} // $\mathbf{E}_{\text{joint}}$ as defined in Eqn.~\ref{jointObjectiveFuncEqn} 
        \State Fix $\mathcal{C}'$, Optimize for $\boldsymbol{\theta}$ using Particle Swarm Method
        \begin{equation}
        \boldsymbol\theta'=\argmin_{\boldsymbol\theta}\|\mathcal{K}(\mathcal{C}',\boldsymbol\theta)-\mathcal{K}(\mathcal{P})_{\text{target}}\|^2.
        \end{equation}
        \State Fix $\boldsymbol{\theta}$, Optimize for $\mathcal{C}'$ using Particle Swam Method
        \begin{equation}
            \tilde{\mathcal{C}}=\argmin_{\mathcal{C}'}\|\mathcal{K}(\mathcal{C}',\boldsymbol\theta)-\mathcal{K}(\mathcal{P})_{\text{target}}\|^2.
        \end{equation}
        \EndWhile
        \State \Return $\tilde{\mathcal{C}}$,$\boldsymbol{\theta}'$
        \EndProcedure
    \end{algorithmic}
\label{jointAlg}
\end{algorithm}
\end{revision}

\subsection{Application to Image-Based Virtual Try-On}
\label{virtualTryOnSec}
This joint material-pose optimization method can be applied directly to image-based virtual try-on. 
We first recover the pose and shape of the human body $<\boldsymbol\theta,\E{z}>$ from the single-view image. 
Then we dress the recovered human body with the reconstructed garments $<\tilde{\mathcal{C}},\tilde{\mathcal{G}},\tilde{\mathbf{U}},\hat{\mathbf{V}}>$ from other images. 
We perform the initial garment registration step (Sec.~\ref{initialGarmentRegistrationSec}) to fit the 3D surface mesh $\hat{\mathbf{V}}$ onto the recovered human body $<\boldsymbol\theta,\E{z}>$.  
%The garment recovered using our method contains the design pattern. 

Existing state-of-the-art virtual try-on rooms require a depth camera for tracking, and overlay the human body with the fitting garment~\cite{ye2014real}.  
Our algorithm, on the other hand, is able to fit the human body from a single 2D image with an optimized virtual outfit recovered from other images. 
We provide the optimized design pattern together with a 3D view of the garment fitted to the human body. 

The fitting step requires iterative optimization in both the garment parameters and the human-body poses. 
As in a real fitting process, we vary the sizing of the outfits for human bodies of different sizes and shapes. 
When editing in parameter space using the methods introduced in the previous section, we ensure that the recovered garment can fit on the human body while minimizing the distortion of the original design.  
For each basic garment, we use one template pattern and the corresponding set of parameters.
To preserve the garment design, we do not change the material properties of the fabric when virtually fitting the recovered garment to a new mannequin. 

\section{Results and Discussion}
\label{sec:results}

We have implemented our algorithm in C++ and demonstrated the effectiveness of our approach throughout the paper. 
In this section, we show example results, performance, and comparisons to other garment recovery methods. 

\subsection{Garment Recovery Results}
\begin{figure}[!hbt]
\centering
\includegraphics[width=0.48\textwidth]{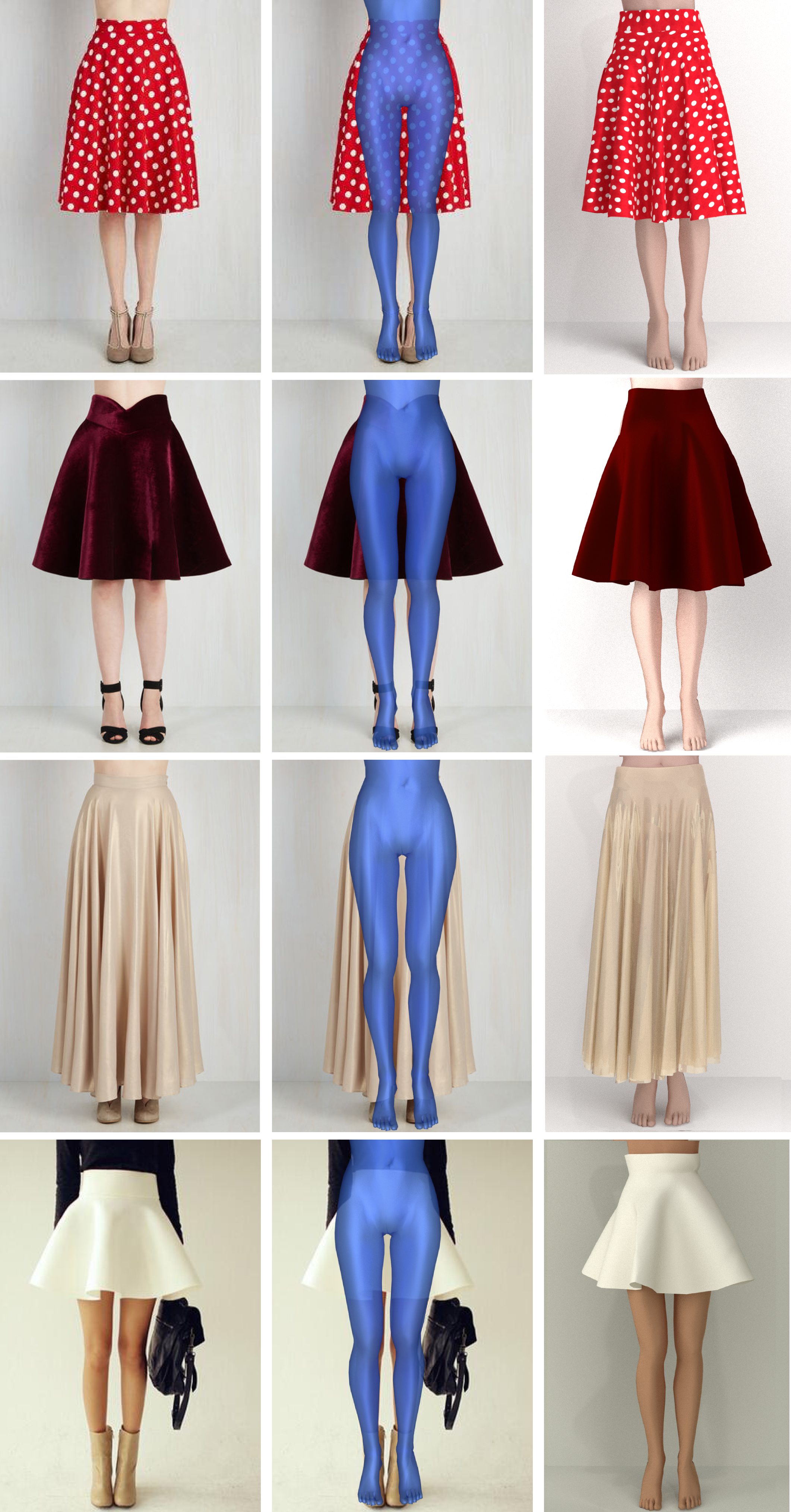}
\caption{{\bf Skirt recovery results.} We recover the partially occluded, folded skirts from the single-view images on the left {\em \protect\cite{skirt,skirtWhiteShort}~$\copyright$}. 
The recovered human body meshes are shown in the middle overlaid on the original images. 
The recovered skirts are shown on the right. }
\label{skirtResultsFig}
\end{figure}

\begin{figure}[!hbt]
\centering
\includegraphics[width=0.48\textwidth]{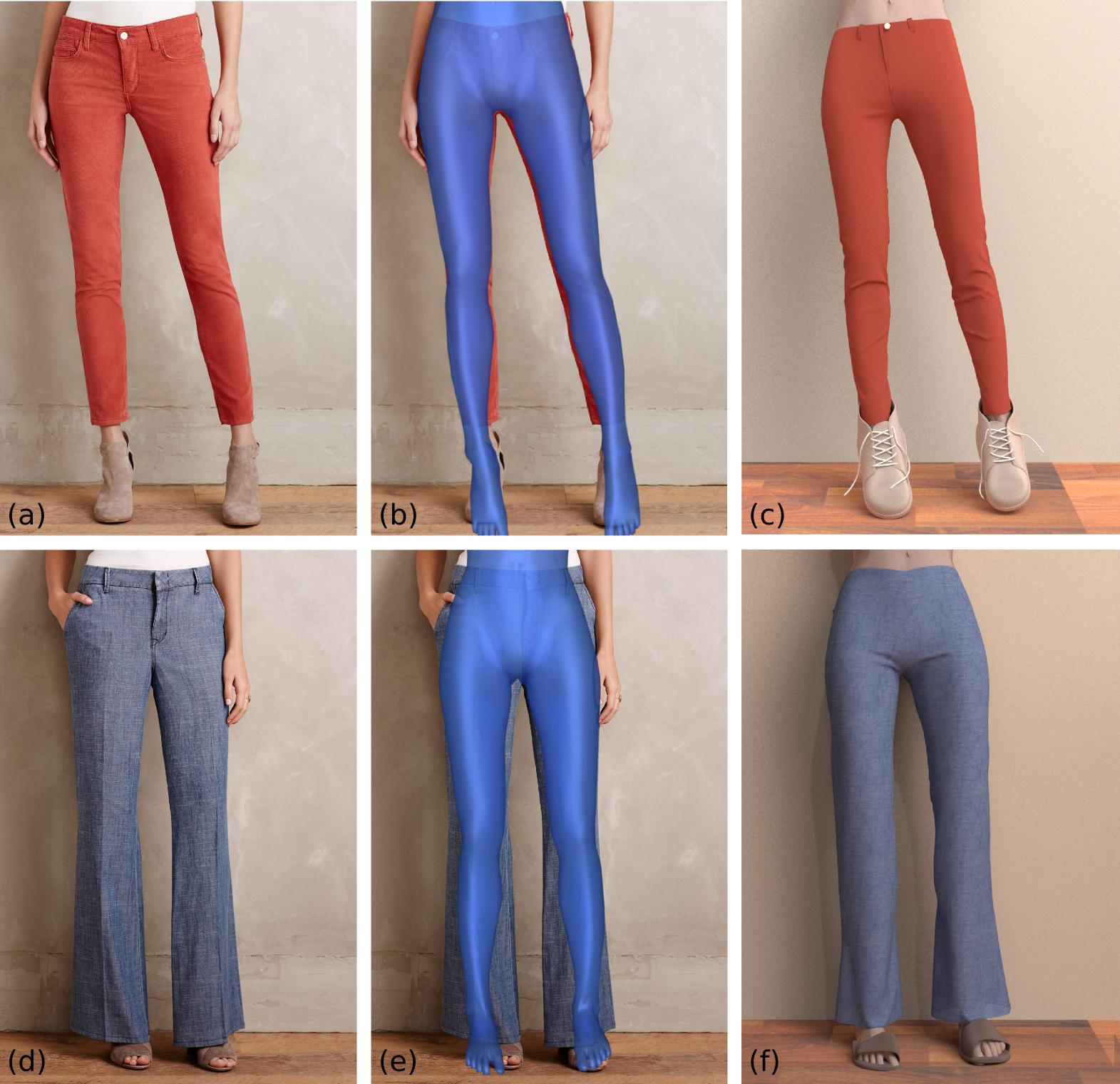}
\vspace{-6mm}
\caption{{\bf Pants recovery results.} The input image (left) {\em \protect\cite{pants}~$\copyright$} and the extracted human body part (middle) and recovered garment (right).}
\label{pantsResultsFig}
\vspace{-3mm}
\end{figure}

\begin{figure}[!hbt]
\centering
\includegraphics[width=0.48\textwidth]{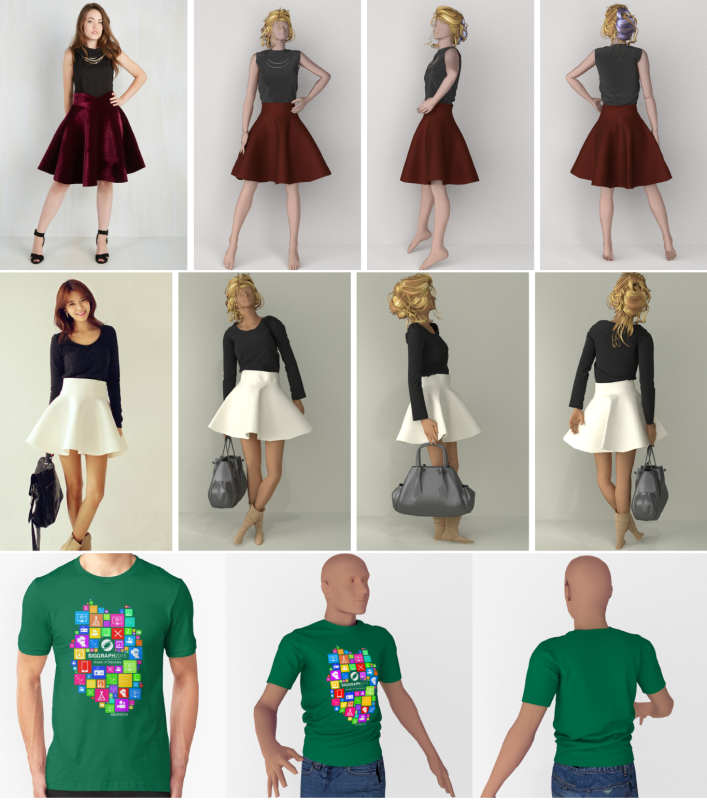}
\vspace{-5mm}
\caption{{\bf Garment recovery results. }For the first two rows, the input image (leftmost) {\em \protect\cite{skirt,skirtWhiteShort,tshirtSIG}~$\copyright$} and recovered garment on the extracted human body.
    In the last row, the input image (leftmost) and the recovered garment on a twisted body.
}
\label{topCombinedFig}
\vspace{-2mm} 
\end{figure}

We show several examples of garment recovery from a single-view image.
In Fig.~\ref{skirtResultsFig} and Fig.~\ref{pantsResultsFig}, we show that our method can recover garments of different styles and materials. 
Fig.~\ref{topCombinedFig} demonstrates the effectiveness of our method for the recovery of partially occluded garments. 
It also shows that our recovered garment can be applied to human bodies in different poses.

\begin{figure}[!hbt]
\centering
\includegraphics[width=0.48\textwidth]{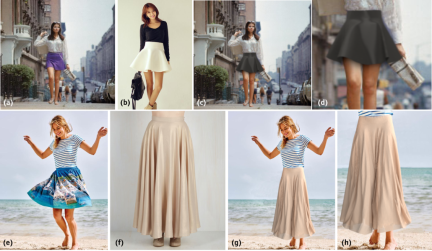}
\vspace{-6mm}
\caption{{\bf Image-based garment transfer results.} We dress the woman in (a) {\em \protect\cite{garmentTransfer1,garmentTransfer2}~$\copyright$} with the skirt we recovered from (b) {\em \protect\cite{skirtWhiteShort,skirt}~$\copyright$}. 
(c) We simulate our recovered skirt with some wind motion to animate the retargeted skirt, 
as shown in (d). 
Another example of garment transfer  is given in (e) - (h).
}
\label{garmentTransfer1Fig}
\end{figure}

{\bf Image-Based Garment Virtual Try-On:}
We show examples of our image-based garment virtual try-on method (Sec.~\ref{virtualTryOnSec}) in  Fig.~\ref{fig:cover} and Fig.~\ref{garmentTransfer1Fig}. 
We can effectively render new outfits onto people from only a single input image. 

\begin{figure}[!hbt]
\centering
\includegraphics[width=0.48\textwidth]{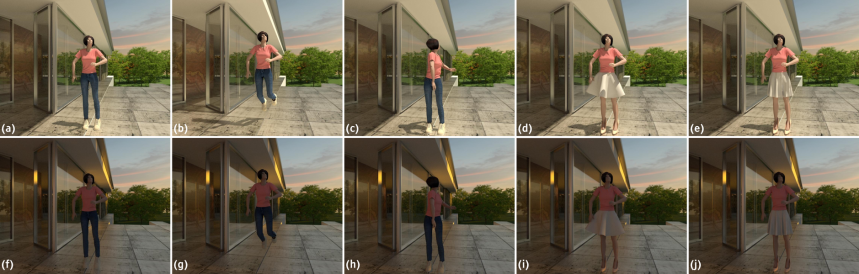}
\vspace{-6mm}
\caption{{\bf Synthetic evaluation scenes.} (a)-(c) fixed body shape with different poses. (d)-(e) fixed body shape with a skirt of different material. 
(f)-(j) same scene setup as (a)-(e) but different lighting condition. }
\label{evaluationFig}
\end{figure}

{\bf Evaluation:}
We evaluate the accuracy of the recovered sizing parameters $\mathcal{G}$ and local curvature $\mathcal{K}$ using synthetic scenes. 
Each synthetic scene has two lighting conditions, mid-day, and sunset (shown in Fig.~\ref{evaluationFig}). 
We fix the both the extrinsic and the intrinsic camera parameters for scene rendering, and the garments are in static equilibrium. 
Through these ten test cases, we can best validate the accuracy and reliability of our method against different body poses and lighting conditions on T-shirts and pants, as the sizing and material parameters are known exactly and do not require noisy measurements and/or data fitting to derive a set of estimated/measured parameters (which are not likely to be $100\%$ accurate) to serve as the ground 
truth.  The evaluation result, 
\begin{revision}
after eliminating the camera scaling factor, is shown in Table~\ref{evalSizingResultsTlb}.
\end{revision}

We found that the lighting conditions mainly affect the body silhouette approximation and the garment folding parsing, 
while the body skeleton approximation is affected by the pose. 
\begin{revision}
Overall, we achieve an accuracy of up to $\mathbf{90.2}\%$ 
for recovering the sizing parameters 
and $\mathbf{82}\%$ for recovering the material parameters
for t-shirts and pants under different body poses and lighting conditions. 
The accuracy is computed as the average accuracy for each parameter from the ground truth.
\end{revision}

\begin{revision}
We evaluate the accuracy of the recovered material properties by measuring the difference between the ground truth and that of the recovered garment for both 
the mean curvature and the material parameters,
as the accuracy of mean-curvature recovery also correlates with
the accuracy of the material-parameter estimation. 
As shown in Table~\ref{evalMatResultsTlb}, 
we are able to achieve an accuracy of
up to $\mathbf{86.7}\%$ and $\mathbf{80.2}\%$, respectively, for 
the recovery of mean curvatures and different material parameters for the skirt. 
\end{revision}

\begin{table}[!ht]
\begin{revision}
%\vspace{-2mm}
%    \centering
%    \resizebox{0.45\textwidth}{!}{
\tbl{{\bf The accuracy of the recovered sizing and material parameters.} The accuracy of the recovered sizing and material parameters of the t-shirt and the pants (in percentages).}{
\begin{tabular}{|l|c|c|c|c|c|c|}
    \hline
    \multicolumn{7}{|c|}{T-Shirt Pants Scene}\\
    \hline
    Pose &a& a&b &b & c & c\\
    \hline
    Lighting &Mid-Day& Sunset&Mid-Day &Sunset &Mid-Day &Sunset\\
    \hline
    $\mathcal{G}_{tshirt}$ Accuracy& $90.2$ & $88.3$ & $89.8$ & $88.1$ & $88.3$ & $86.3$  \\
    \hline
    $\mathcal{G}_{pants}$ Accuracy& $89.3$ & $87.6$ & $85.8$ & $83.3$ & $88.2$ & $87.5$  \\
    \hline
    $\mathcal{C}_{tshirt}$ Accuracy& $80.6$ & $81.3$ & $79.2$ & $81.5$ & $80.9$ & $82.0$  \\
    \hline
    $\mathcal{C}_{pants}$ Accuracy& $80.3$ & $78.6$ & $80.0$ & $80.7$ & $80.3$ & $80.5$  \\
    \hline
\end{tabular}}
\label{evalSizingResultsTlb}
\end{revision}
\end{table}

\begin{table}[!ht]
\begin{revision}
    %\centering
    %\resizebox{0.45\textwidth}{!}{
\tbl{{\bf The accuracy of recovered garment curvature and material parameters.} The accuracy of the recovered garment local mean curvature and material parameters of the skirt (in percentages).}{
\begin{tabular}{|l|c|c|c|c|}
    \hline
    \multicolumn{5}{|c|}{T-Shirt Skirt Scene}\\
    \hline
    Material &Low Bending& Low Bending& High Bending & High Bending \\
    \hline
    Lighting &Mid-Day& Sunset&Mid-Day &Sunset \\
    \hline
    $\mathcal{K}_{skirt}$ Accuracy& $86.7$ & $83.4$ & $85.3$ & $82.5$  \\
    \hline
    $\mathcal{C}_{skirt}$ Accuracy& $80.2$ & $78.9$ & $81.3$ & $78.3$  \\
    \hline
\end{tabular}}
\label{evalMatResultsTlb}
\end{revision}
\end{table}

\subsection{Comparison with Other Related Work}

\begin{figure}[!hbt]
\centering
\includegraphics[width=0.48\textwidth]{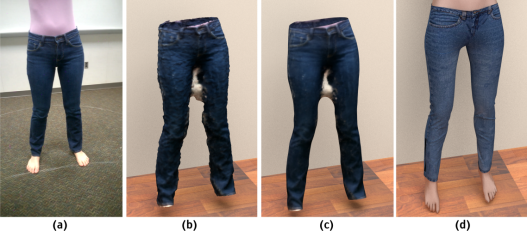}
\vspace{-6mm}
\caption{{\bf Comparison.} (a) One frame of the video along with (b) the CMP-MVS results before and (c) after smoothing. 
(d) Our results using only one frame of the video. }
\label{pantsComparisonFig}
\vspace{-4mm}
\end{figure}

\begin{figure}[!hbt]
\centering
\includegraphics[scale=0.38]{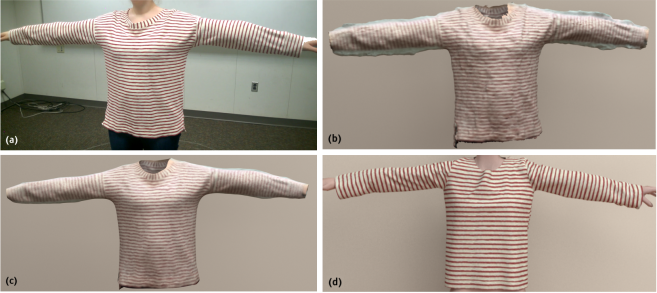}
\vspace{-3mm}
\caption{{\bf Comparison.} (a) One frame of the multi-view video along with (b) the CMP-MVS results before and (c) after smoothing. 
(d) Our results using only one frame of the video.}
\label{tshirtComparisonFig}
\vspace{-2mm}
\end{figure}

We compare our results with the multi-view reconstruction method CMP-MVS~\cite{Jancosek2011Multiview} together with the structure-from-motion framework~\cite{wu2011visualsfm,Wu2013Towards}.
For a fair comparison, we apply smoothing~\cite{Taubin1995Signal} to the results of their work.
Fig.~\ref{pantsComparisonFig} and Fig.~\ref{tshirtComparisonFig} show that the garment recovered using our method is clean and comparable in visual quality to the recovered garments using multi-view 
methods.  In addition, we are able to estimate the material properties from one single-view
image for virtual try-on applications.

\begin{revision}
We further compare the results of our work against two recent methods -- one 
using 3D depth information and an expert-designed 3D database~\cite{chen2015garment}, and the other using a large database of manually labeled garment 
images~\cite{jeong2015garment}.
Our method, which does not require depth information, an expert-designed 3D database, or a large manually labeled garment image database, achieves a comparable level of high accuracy to Chen et al.~\shortcite{chen2015garment} (see Fig.~15)
% \ref{fig:dressCompareFig} 
and higher visual quality when compared with Jeong et al.~\shortcite{jeong2015garment} 
(see Fig.~16).
% \ref{fig:shortsCompareFig}).  
In addition, our method is able to recover material and estimate sizing parameters 
directly from a given image.
\end{revision}

\begin{figure}[!hbt]
\centering
\includegraphics[scale=0.13]{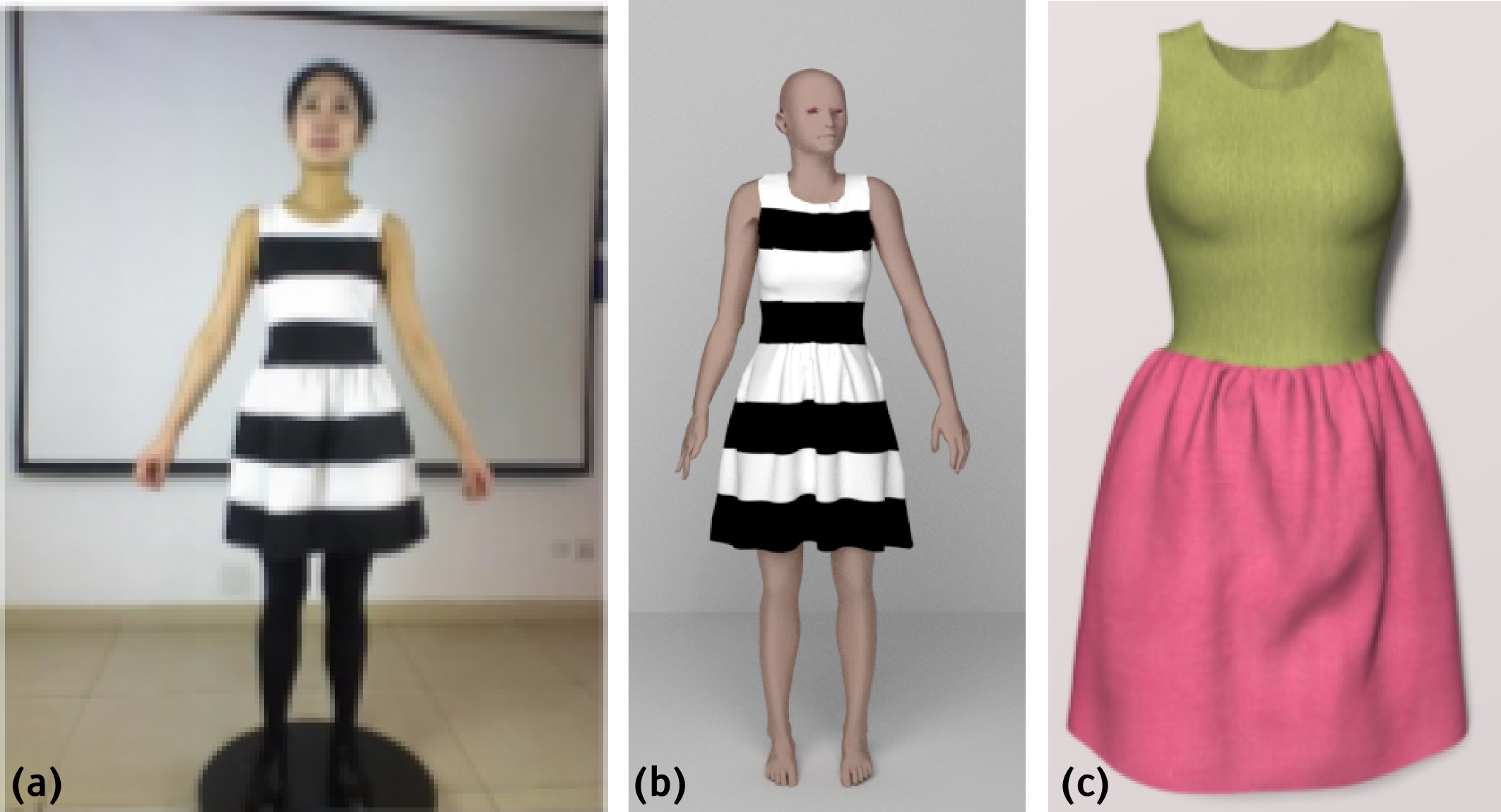}
\vspace{-3mm}
\begin{revision}
\caption{{\bf Comparison.} (a) input image (\copyright~~2015 ACM) from paper Chen et al.~\shortcite{chen2015garment} Figure 12. (b) our garment recover results from only a single-view RGB image (a). (c) recovery results (\copyright~~2015 ACM) from Chen et al.~\shortcite{chen2015garment} using both RGB image and depth information. }
\end{revision}
\label{fig:dressCompareFig}
\vspace{-2mm}
\end{figure}

\begin{figure}[!hbt]
\centering
\includegraphics[scale=0.13]{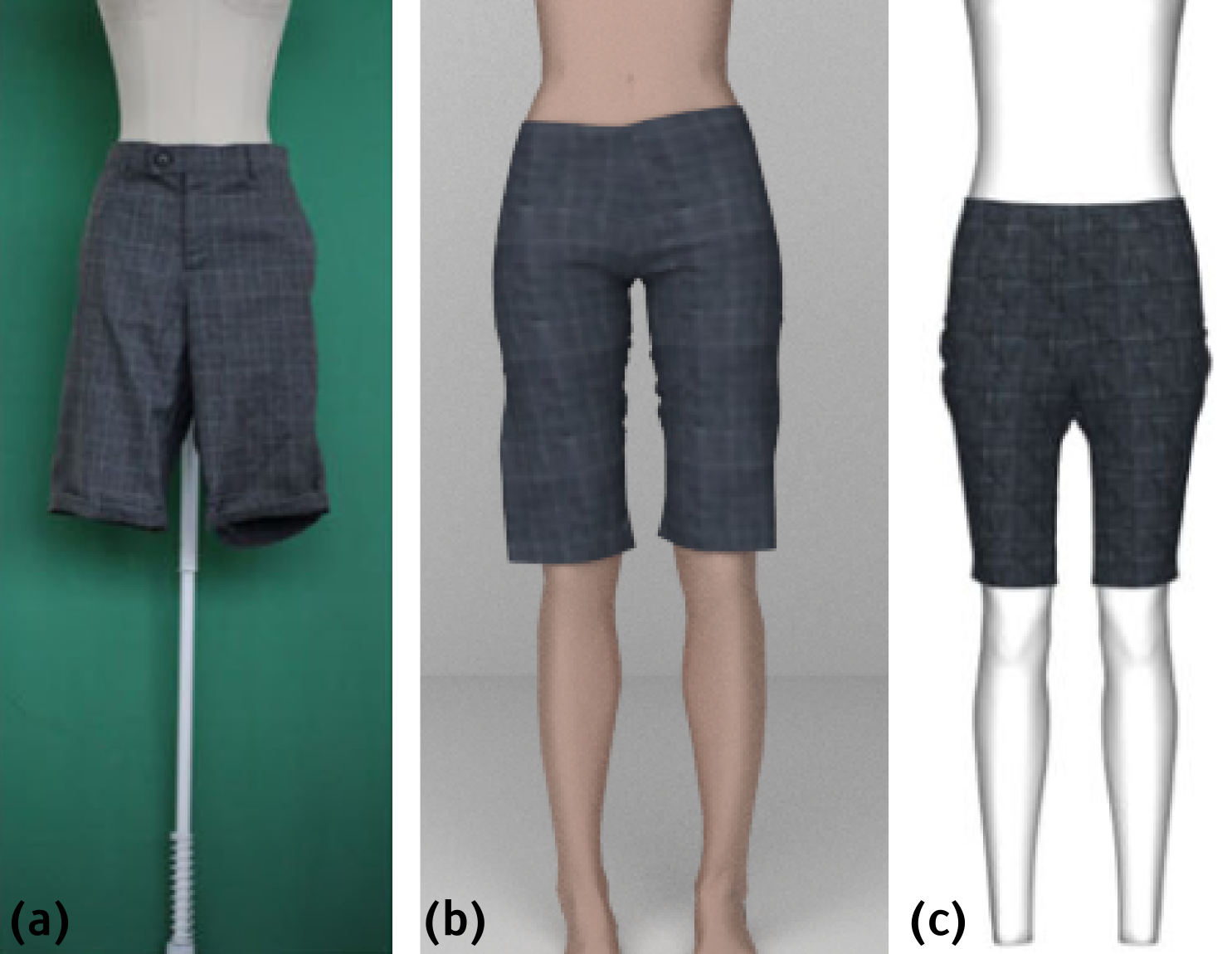}
\vspace{-3mm}
\begin{revision}
\caption{{\bf Comparison.} (a) input image (\copyright~~2015 Wiley) from Figure 3 in Jeong et al.~\shortcite{jeong2015garment}. (b) our garment recover results from (a). (c) recovery results 
(\copyright~~2015 Wiley) from Jeong et al.~\shortcite{jeong2015garment}. }
\end{revision}
\label{fig:shortsCompareFig}
\vspace{-2mm}
\end{figure}

\subsection{Performance}

We run our method on a desktop with an Intel(R) Core(TM) i7 CPU, 3.20GHz. 
For each garment, our pipeline takes on average 4 to 6 hours.
The garment parameter identification (Sec.~\ref{garmentParameterSec}) and joint material-pose optimization (Sec.~\ref{jointOptimizationSec}) takes around $60\%-80\%$ of the entire process. 
The preprocessing step (Sec.~\ref{preprocessingSec}) takes around $20\%-30\%$. 
The performance depends largely on the complexity of the garment, the image quality, and how much the garment is occluded. 

\begin{revision}
\subsection{Discussions}

\noindent
{\bf Estimation of Material Parameters: }
Our material recovery method depends on the existence of wrinkles and folds 
of the garment.
In cases where no or very few wrinkles or folds are present, other image features, 
such as textures and shading, would be required to identify the material
properties.
In most garments like shirts, skirts, or dresses, wrinkles and folds 
are common (especially around the joints or from the garment stylization),
and can be highly informative with regards to garment material properties.
Based on this observation, we are able to estimate {\em material
parameters} as well as recover garment geometry from single-view images. 
This capability is one of our main objectives, and it is the key feature 
differentiating our work from existing techniques.

\noindent
{\bf Accuracy of Geometry Reconstruction: }
In general, it is expected that recovery from single-view images should yield 
less accurate results than from standard 3D reconstruction and/or the most recent 
3D multi-view methods.  Our method adopts accurate physics-based 
cloth simulation to assist in the recovery process and achieves comparable 
visual quality, with a focus on capturing plausible wrinkles and folds, as
well as material parameters required for virtual try-on using only photographs.

However, it is important to note that high visual quality does not always 
guarantee geometric accuracy in 3D garment recovery. At the same time, for 
some applications such as virtual try-on, rapid design, and prototyping, 
it is unclear if a high degree of geometric accuracy is required; 
it is also unknown how much error tolerance is needed for the comfortable 
fitting of garments.  These are important considerations for further 
investigation in application to fashion design and e-commerce. 
\end{revision}

\subsection{Limitations}

The current implementation of our approach depends on two databases: 
a database of commonly available garment templates and a database of
human-body models. 

%The garment template is chosen from the garment template database. 
The range of garments we can recover is, to some extent, limited by the available garment templates. 
Our parameter identification method can only generate garments that are ``morphable'' from the garment template, i.e. homeomorphic to the garment template.
For example, since we use only one template for each garment type, we cannot yet model variations in some clothing details, e.g. multi-layered skirts, or collars on shirts.  But for those garments that are not morphable from the template, our method can recover whichever version of the garment is closest to the actual garment. With a more extensive set of templates, we can begin to model more variations of styles and cuts, with richer garment details. 

Another limitation is the human body shape recovery. 
Our reduced human body shape is described by a set of semantic parameters $\mathbf{z}$. 
The representation of this set of semantic parameters is not enormous, though it is sufficient to include most of the common human body shapes, as shown in our images. 
\begin{revision}
The known artifacts of linear human shape blending can also affect results. 
Aside from the human body shape recovery, our method is also limited by the 
state-of-art 3D human pose recovery methods. 
Manual intervention is needed when these methods fail to output a reasonably
accurate 3D human pose.
\end{revision}

\section{Conclusion and Future Work}
In this paper we present an algorithm for highly detailed garment recovery from a single-view image. Our approach recovers a 3D mesh of the garment together with the 2D design pattern, fine wrinkles and folds, and material parameters.
The recovered garment can be re-targeted to other human bodies of different shapes, sizes, and poses for virtual try-on and character animation.

In addition to addressing some limitations mentioned above, there are many possible future research directions.   
First of all, we plan to develop a parallelized implementation of our system 
on GPU or a many-core CPU for fast garment recovery.
Both the underlying cloth simulator and the optimization process can be significantly accelerated.  
We also plan to extend our approach to enable fabric material transfer from videos for interactive 
virtual try-on.  Furthermore, we hope to explore possible perception metrics, similar in spirit 
to~\cite{sigal2015perceptual}.

%\input{wrinkle2D}
%\input{wrinkle3D}
% \section*{Acknowledgements}
\bibliographystyle{acmtog}
\bibliography{bibs/main}

\begin{thebibliography}{}

\bibitem[\protect\citeauthoryear{Agarwal and Triggs}{Agarwal and
  Triggs}{2006}]{agarwal2006recovering}
{\sc Agarwal, A.} {\sc and} {\sc Triggs, B.} 2006.
\newblock Recovering 3d human pose from monocular images.
\newblock {\em Pattern Analysis and Machine Intelligence, IEEE Transactions
  on\/}~{\em 28,\/}~1, 44--58.

\bibitem[\protect\citeauthoryear{AliExpress}{AliExpress}{2015}]{skirtWhiteShor%
t}
{\sc AliExpress}. 2015.
\newblock \url{http://www.aliexpress.com}.

\bibitem[\protect\citeauthoryear{Anguelov, Srinivasan, Koller, Thrun, Rodgers,
  and Davis}{Anguelov et~al\mbox{.}}{2005}]{anguelov2005scape}
{\sc Anguelov, D.}, {\sc Srinivasan, P.}, {\sc Koller, D.}, {\sc Thrun, S.},
  {\sc Rodgers, J.}, {\sc and} {\sc Davis, J.} 2005.
\newblock Scape: shape completion and animation of people.
\newblock In {\em ACM Transactions on Graphics (TOG)}. Vol.~24. ACM, 408--416.

\bibitem[\protect\citeauthoryear{Anthropologie}{Anthropologie}{2015}]{pants}
{\sc Anthropologie}. 2015.
\newblock \url{http://www.anthropologie.com}.

\bibitem[\protect\citeauthoryear{Balan, Sigal, Black, Davis, and
  Haussecker}{Balan et~al\mbox{.}}{2007}]{balan2007detailed}
{\sc Balan, A.~O.}, {\sc Sigal, L.}, {\sc Black, M.~J.}, {\sc Davis, J.~E.},
  {\sc and} {\sc Haussecker, H.~W.} 2007.
\newblock Detailed human shape and pose from images.
\newblock In {\em 2007 IEEE Conference on Computer Vision and Pattern
  Recognition}. IEEE, 1--8.

\bibitem[\protect\citeauthoryear{Baraff and Witkin}{Baraff and
  Witkin}{1998}]{baraff1998large}
{\sc Baraff, D.} {\sc and} {\sc Witkin, A.} 1998.
\newblock Large steps in cloth simulation.
\newblock In {\em Proceedings of the 25th annual conference on Computer
  graphics and interactive techniques}. ACM, 43--54.

\bibitem[\protect\citeauthoryear{Barnfield}{Barnfield}{2012}]{barnfield2012the}
{\sc Barnfield, J.} 2012.
\newblock {\em The pattern making primer : all you need to know about
  designing, adapting and customizing sewing patterns}.
\newblock Barron's Educational Series, Inc, Hauppauge, N.Y.

\bibitem[\protect\citeauthoryear{B{\'e}rard, Bradley, Nitti, Beeler, and
  Gross}{B{\'e}rard et~al\mbox{.}}{2014}]{berard2014high}
{\sc B{\'e}rard, P.}, {\sc Bradley, D.}, {\sc Nitti, M.}, {\sc Beeler, T.},
  {\sc and} {\sc Gross, M.} 2014.
\newblock High-quality capture of eyes.
\newblock {\em ACM Transactions on Graphics (TOG)\/}~{\em 33,\/}~6, 223.

\bibitem[\protect\citeauthoryear{Berthouzoz, Garg, Kaufman, Grinspun, and
  Agrawala}{Berthouzoz et~al\mbox{.}}{2013}]{berthouzoz2013parsing}
{\sc Berthouzoz, F.}, {\sc Garg, A.}, {\sc Kaufman, D.~M.}, {\sc Grinspun, E.},
  {\sc and} {\sc Agrawala, M.} 2013.
\newblock Parsing sewing patterns into 3d garments.
\newblock {\em ACM Transactions on Graphics (TOG)\/}~{\em 32,\/}~4, 85.

\bibitem[\protect\citeauthoryear{Boden}{Boden}{2015}]{garmentTransfer2}
{\sc Boden}. 2015.
\newblock \url{http://www.bodenusa.com}.

\bibitem[\protect\citeauthoryear{Bridson, Fedkiw, and Anderson}{Bridson
  et~al\mbox{.}}{2002}]{bridson2002robust}
{\sc Bridson, R.}, {\sc Fedkiw, R.}, {\sc and} {\sc Anderson, J.} 2002.
\newblock Robust treatment of collisions, contact and friction for cloth
  animation.
\newblock In {\em ACM Transactions on Graphics (ToG)}. Vol.~21. ACM, 594--603.

\bibitem[\protect\citeauthoryear{Bridson, Marino, and Fedkiw}{Bridson
  et~al\mbox{.}}{2003}]{bridson2003simulation}
{\sc Bridson, R.}, {\sc Marino, S.}, {\sc and} {\sc Fedkiw, R.} 2003.
\newblock Simulation of clothing with folds and wrinkles.
\newblock In {\em Proceedings of the 2003 ACM SIGGRAPH/Eurographics symposium
  on Computer animation}. Eurographics Association, 28--36.

\bibitem[\protect\citeauthoryear{Bronstein, Bronstein, and Kimmel}{Bronstein
  et~al\mbox{.}}{2006}]{bronstein2006efficient}
{\sc Bronstein, A.~M.}, {\sc Bronstein, M.~M.}, {\sc and} {\sc Kimmel, R.}
  2006.
\newblock Efficient computation of isometry-invariant distances between
  surfaces.
\newblock {\em SIAM Journal on Scientific Computing\/}~{\em 28,\/}~5,
  1812--1836.

\bibitem[\protect\citeauthoryear{Bronstein, Bronstein, and Kimmel}{Bronstein
  et~al\mbox{.}}{2008}]{bronstein2008numerical}
{\sc Bronstein, A.~M.}, {\sc Bronstein, M.~M.}, {\sc and} {\sc Kimmel, R.}
  2008.
\newblock {\em Numerical geometry of non-rigid shapes}.
\newblock Springer Science \& Business Media.

\bibitem[\protect\citeauthoryear{Brouet, Sheffer, Boissieux, and Cani}{Brouet
  et~al\mbox{.}}{2012}]{brouet2012design}
{\sc Brouet, R.}, {\sc Sheffer, A.}, {\sc Boissieux, L.}, {\sc and} {\sc Cani,
  M.-P.} 2012.
\newblock Design preserving garment transfer.
\newblock {\em ACM Trans. Graph.\/}~{\em 31,\/}~4, 36.

\bibitem[\protect\citeauthoryear{Cao, Weng, Lin, and Zhou}{Cao
  et~al\mbox{.}}{2013}]{cao20133d}
{\sc Cao, C.}, {\sc Weng, Y.}, {\sc Lin, S.}, {\sc and} {\sc Zhou, K.} 2013.
\newblock 3d shape regression for real-time facial animation.
\newblock {\em ACM Transactions on Graphics (TOG)\/}~{\em 32,\/}~4, 41.

\bibitem[\protect\citeauthoryear{Casati, Daviet, and Bertails-Descoubes}{Casati
  et~al\mbox{.}}{2016}]{casati2016inverse}
{\sc Casati, R.}, {\sc Daviet, G.}, {\sc and} {\sc Bertails-Descoubes, F.}
  2016.
\newblock Inverse elastic cloth design with contact and friction.
\newblock Ph.D. thesis, Inria Grenoble Rh{\^o}ne-Alpes, Universit{\'e} de
  Grenoble.

\bibitem[\protect\citeauthoryear{Chai, Wang, Weng, Yu, Guo, and Zhou}{Chai
  et~al\mbox{.}}{2012}]{chai2012single}
{\sc Chai, M.}, {\sc Wang, L.}, {\sc Weng, Y.}, {\sc Yu, Y.}, {\sc Guo, B.},
  {\sc and} {\sc Zhou, K.} 2012.
\newblock Single-view hair modeling for portrait manipulation.
\newblock {\em ACM Trans. Graph.\/}~{\em 31,\/}~4 (July), 116:1--116:8.

\bibitem[\protect\citeauthoryear{Chen, Guo, Zhou, and Zhao}{Chen
  et~al\mbox{.}}{2013}]{chen2013deformable}
{\sc Chen, X.}, {\sc Guo, Y.}, {\sc Zhou, B.}, {\sc and} {\sc Zhao, Q.} 2013.
\newblock Deformable model for estimating clothed and naked human shapes from a
  single image.
\newblock {\em The Visual Computer\/}~{\em 29,\/}~11, 1187--1196.

\bibitem[\protect\citeauthoryear{Chen, Zhou, Lu, Wang, Bi, and Tan}{Chen
  et~al\mbox{.}}{2015}]{chen2015garment}
{\sc Chen, X.}, {\sc Zhou, B.}, {\sc Lu, F.}, {\sc Wang, L.}, {\sc Bi, L.},
  {\sc and} {\sc Tan, P.} 2015.
\newblock Garment modeling with a depth camera.
\newblock {\em ACM Transactions on Graphics (TOG)\/}~{\em 34,\/}~6, 203.

\bibitem[\protect\citeauthoryear{Curtis, Tamstorf, and Manocha}{Curtis
  et~al\mbox{.}}{2008}]{curtis2008fast}
{\sc Curtis, S.}, {\sc Tamstorf, R.}, {\sc and} {\sc Manocha, D.} 2008.
\newblock Fast collision detection for deformable models using
  representative-triangles.
\newblock In {\em Proceedings of the 2008 symposium on Interactive 3D graphics
  and games}. ACM, 61--69.

\bibitem[\protect\citeauthoryear{Decaudin, Julius, Wither, Boissieux, Sheffer,
  and Cani}{Decaudin et~al\mbox{.}}{2006}]{decaudin2006virtual}
{\sc Decaudin, P.}, {\sc Julius, D.}, {\sc Wither, J.}, {\sc Boissieux, L.},
  {\sc Sheffer, A.}, {\sc and} {\sc Cani, M.-P.} 2006.
\newblock Virtual garments: A fully geometric approach for clothing design.
\newblock In {\em Computer Graphics Forum}. Vol.~25. Wiley Online Library,
  625--634.

\bibitem[\protect\citeauthoryear{English and Bridson}{English and
  Bridson}{2008}]{english2008animating}
{\sc English, E.} {\sc and} {\sc Bridson, R.} 2008.
\newblock Animating developable surfaces using nonconforming elements.
\newblock In {\em ACM Transactions on Graphics (TOG)}. Vol.~27. ACM, 66.

\bibitem[\protect\citeauthoryear{Farabet, Couprie, Najman, and LeCun}{Farabet
  et~al\mbox{.}}{2013}]{Farabet13}
{\sc Farabet, C.}, {\sc Couprie, C.}, {\sc Najman, L.}, {\sc and} {\sc LeCun,
  Y.} 2013.
\newblock Learning hierarchical features for scene labeling.
\newblock In {\em Pattern Analysis and Machine Intelligence}.

\bibitem[\protect\citeauthoryear{FashionableShoes}{FashionableShoes}{2013}]{ga%
rmentTransfer1}
{\sc FashionableShoes}. 2013.
\newblock \url{http://bestfashionableshoess.blogspot.com}.

\bibitem[\protect\citeauthoryear{FashionUnited}{FashionUnited}{2016}]{fashion_%
stat}
{\sc FashionUnited}. 2016.
\newblock Global fashion industry statistics - international apparel.

\bibitem[\protect\citeauthoryear{Goldenthal, Harmon, Fattal, Bercovier, and
  Grinspun}{Goldenthal et~al\mbox{.}}{2007}]{goldenthal2007efficient}
{\sc Goldenthal, R.}, {\sc Harmon, D.}, {\sc Fattal, R.}, {\sc Bercovier, M.},
  {\sc and} {\sc Grinspun, E.} 2007.
\newblock Efficient simulation of inextensible cloth.
\newblock {\em ACM Transactions on Graphics (TOG)\/}~{\em 26,\/}~3, 49.

\bibitem[\protect\citeauthoryear{Govindaraju, Kabul, Lin, and
  Manocha}{Govindaraju et~al\mbox{.}}{2007}]{govindaraju2007fast}
{\sc Govindaraju, N.~K.}, {\sc Kabul, I.}, {\sc Lin, M.~C.}, {\sc and} {\sc
  Manocha, D.} 2007.
\newblock Fast continuous collision detection among deformable models using
  graphics processors.
\newblock {\em Computers \& Graphics\/}~{\em 31,\/}~1, 5--14.

\bibitem[\protect\citeauthoryear{Hasler, Asbach, Rosenhahn, Ohm, and
  Seidel}{Hasler et~al\mbox{.}}{2006}]{hasler2006physically}
{\sc Hasler, N.}, {\sc Asbach, M.}, {\sc Rosenhahn, B.}, {\sc Ohm, J.-R.}, {\sc
  and} {\sc Seidel, H.-P.} 2006.
\newblock Physically based tracking of cloth.
\newblock In {\em Proc. of the International Workshop on Vision, Modeling, and
  Visualization, VMV}. 49--56.

\bibitem[\protect\citeauthoryear{Hasler, Stoll, Sunkel, Rosenhahn, and
  Seidel}{Hasler et~al\mbox{.}}{2009}]{hasler2009statistical}
{\sc Hasler, N.}, {\sc Stoll, C.}, {\sc Sunkel, M.}, {\sc Rosenhahn, B.}, {\sc
  and} {\sc Seidel, H.-P.} 2009.
\newblock A statistical model of human pose and body shape.
\newblock In {\em Computer Graphics Forum}. Vol.~28. Wiley Online Library,
  337--346.

\bibitem[\protect\citeauthoryear{Hillsweddingdress}{Hillsweddingdress}{2015}]{%
garmentTransfer5}
{\sc Hillsweddingdress}. 2015.
\newblock \url{http://hillsweddingdress.xyz}.

\bibitem[\protect\citeauthoryear{House and Breen}{House and
  Breen}{2000}]{house2000cloth}
{\sc House, D.~H.} {\sc and} {\sc Breen, D.~E.} 2000.
\newblock {\em Cloth modeling and animation}.
\newblock AK Peters.

\bibitem[\protect\citeauthoryear{Igarashi, Moscovich, and Hughes}{Igarashi
  et~al\mbox{.}}{2005}]{igarashi2005rigid}
{\sc Igarashi, T.}, {\sc Moscovich, T.}, {\sc and} {\sc Hughes, J.~F.} 2005.
\newblock As-rigid-as-possible shape manipulation.
\newblock In {\em ACM transactions on Graphics (TOG)}. Vol.~24. ACM,
  1134--1141.

\bibitem[\protect\citeauthoryear{Jancosek and Pajdla}{Jancosek and
  Pajdla}{2011}]{Jancosek2011Multiview}
{\sc Jancosek, M.} {\sc and} {\sc Pajdla, T.} 2011.
\newblock Multi-view reconstruction preserving weakly-supported surfaces.
\newblock In {\em Computer Vision and Pattern Recognition (CVPR), 2011 IEEE
  Conference on}. 3121--3128.

\bibitem[\protect\citeauthoryear{Jeong, Han, and Ko}{Jeong
  et~al\mbox{.}}{2015}]{jeong2015garment}
{\sc Jeong, M.-H.}, {\sc Han, D.-H.}, {\sc and} {\sc Ko, H.-S.} 2015.
\newblock Garment capture from a photograph.
\newblock {\em Computer Animation and Virtual Worlds\/}~{\em 26,\/}~3-4,
  291--300.

\bibitem[\protect\citeauthoryear{Kavan, Sloan, and O'Sullivan}{Kavan
  et~al\mbox{.}}{2010}]{kavan2010fast}
{\sc Kavan, L.}, {\sc Sloan, P.-P.}, {\sc and} {\sc O'Sullivan, C.} 2010.
\newblock Fast and efficient skinning of animated meshes.
\newblock In {\em Computer Graphics Forum}. Vol.~29. Wiley Online Library,
  327--336.

\bibitem[\protect\citeauthoryear{Kennedy}{Kennedy}{2010}]{kennedy2010particle}
{\sc Kennedy, J.} 2010.
\newblock Particle swarm optimization.
\newblock In {\em Encyclopedia of Machine Learning}. Springer, 760--766.

\bibitem[\protect\citeauthoryear{Levin, Lischinski, and Weiss}{Levin
  et~al\mbox{.}}{2008}]{levin2008closed}
{\sc Levin, A.}, {\sc Lischinski, D.}, {\sc and} {\sc Weiss, Y.} 2008.
\newblock A closed-form solution to natural image matting.
\newblock {\em Pattern Analysis and Machine Intelligence, IEEE Transactions
  on\/}~{\em 30,\/}~2, 228--242.

\bibitem[\protect\citeauthoryear{Li, Luo, Vlasic, Peers, Popovi\'{c}, Pauly,
  and Rusinkiewicz}{Li et~al\mbox{.}}{2012}]{li2012temporally}
{\sc Li, H.}, {\sc Luo, L.}, {\sc Vlasic, D.}, {\sc Peers, P.}, {\sc
  Popovi\'{c}, J.}, {\sc Pauly, M.}, {\sc and} {\sc Rusinkiewicz, S.} 2012.
\newblock Temporally coherent completion of dynamic shapes.
\newblock {\em ACM Transactions on Graphics\/}~{\em 31,\/}~1 (January).

\bibitem[\protect\citeauthoryear{Li, Sumner, and Pauly}{Li
  et~al\mbox{.}}{2008}]{li2008global}
{\sc Li, H.}, {\sc Sumner, R.~W.}, {\sc and} {\sc Pauly, M.} 2008.
\newblock Global correspondence optimization for non-rigid registration of
  depth scans.
\newblock In {\em Computer graphics forum}. Vol.~27. Wiley Online Library,
  1421--1430.

\bibitem[\protect\citeauthoryear{Li, Sun, Tang, and Shum}{Li
  et~al\mbox{.}}{2004}]{li2004lazy}
{\sc Li, Y.}, {\sc Sun, J.}, {\sc Tang, C.-K.}, {\sc and} {\sc Shum, H.-Y.}
  2004.
\newblock Lazy snapping.
\newblock In {\em ACM Transactions on Graphics (ToG)}. Vol.~23. ACM, 303--308.

\bibitem[\protect\citeauthoryear{Lindner}{Lindner}{2015}]{ecommerce_stat}
{\sc Lindner, M.} 2015.
\newblock Global e-commerce sales set to grow 25\% in 2015.
\newblock
  https://www.internetretailer.com/2015/07/29/global-e-commerce-set-grow-25-20%
15.

\bibitem[\protect\citeauthoryear{Liu and Nocedal}{Liu and
  Nocedal}{1989}]{liu1989limited}
{\sc Liu, D.~C.} {\sc and} {\sc Nocedal, J.} 1989.
\newblock On the limited memory bfgs method for large scale optimization.
\newblock {\em Mathematical programming\/}~{\em 45,\/}~1-3, 503--528.

\bibitem[\protect\citeauthoryear{Long, Shelhamer, and Darrell}{Long
  et~al\mbox{.}}{2015}]{long_shelhamer_fcn}
{\sc Long, J.}, {\sc Shelhamer, E.}, {\sc and} {\sc Darrell, T.} 2015.
\newblock Fully convolutional networks for semantic segmentation.
\newblock {\em CVPR (to appear)\/}.

\bibitem[\protect\citeauthoryear{Meng, Wang, and Jin}{Meng
  et~al\mbox{.}}{2012}]{meng2012flexible}
{\sc Meng, Y.}, {\sc Wang, C.~C.}, {\sc and} {\sc Jin, X.} 2012.
\newblock Flexible shape control for automatic resizing of apparel products.
\newblock {\em Computer-Aided Design\/}~{\em 44,\/}~1, 68--76.

\bibitem[\protect\citeauthoryear{ModCloth}{ModCloth}{2015}]{skirt}
{\sc ModCloth}. 2015.
\newblock \url{http://www.modcloth.com}.

\bibitem[\protect\citeauthoryear{Moeslund, Hilton, and Kr{\"u}ger}{Moeslund
  et~al\mbox{.}}{2006}]{moeslund2006survey}
{\sc Moeslund, T.~B.}, {\sc Hilton, A.}, {\sc and} {\sc Kr{\"u}ger, V.} 2006.
\newblock A survey of advances in vision-based human motion capture and
  analysis.
\newblock {\em Computer vision and image understanding\/}~{\em 104,\/}~2,
  90--126.

\bibitem[\protect\citeauthoryear{Nagano, Fyffe, Alexander, Barbi\c{c}, Li,
  Ghosh, and Debevec}{Nagano
  et~al\mbox{.}}{2015}]{Nagano:2015:SMD:2809654.2766894}
{\sc Nagano, K.}, {\sc Fyffe, G.}, {\sc Alexander, O.}, {\sc Barbi\c{c}, J.},
  {\sc Li, H.}, {\sc Ghosh, A.}, {\sc and} {\sc Debevec, P.} 2015.
\newblock Skin microstructure deformation with displacement map convolution.
\newblock {\em ACM Trans. Graph.\/}~{\em 34,\/}~4 (July), 109:1--109:10.

\bibitem[\protect\citeauthoryear{Narain, Samii, and O'Brien}{Narain
  et~al\mbox{.}}{2012}]{narain2012adaptive}
{\sc Narain, R.}, {\sc Samii, A.}, {\sc and} {\sc O'Brien, J.~F.} 2012.
\newblock Adaptive anisotropic remeshing for cloth simulation.
\newblock {\em ACM Transactions on Graphics (TOG)\/}~{\em 31,\/}~6, 152.

\bibitem[\protect\citeauthoryear{Ng and Grimsdale}{Ng and
  Grimsdale}{1996}]{ng1996computer}
{\sc Ng, H.~N.} {\sc and} {\sc Grimsdale, R.~L.} 1996.
\newblock Computer graphics techniques for modeling cloth.
\newblock {\em Computer Graphics and Applications, IEEE\/}~{\em 16,\/}~5,
  28--41.

\bibitem[\protect\citeauthoryear{Pinheiro and Collobert}{Pinheiro and
  Collobert}{2014}]{Pinheiro14}
{\sc Pinheiro, P.~H.} {\sc and} {\sc Collobert, R.} 2014.
\newblock Recurrent convolutional neural networks for scene labeling.
\newblock In {\em ICML}.

\bibitem[\protect\citeauthoryear{Popa, Zhou, Bradley, Kraevoy, Fu, Sheffer, and
  Heidrich}{Popa et~al\mbox{.}}{2009}]{popa2009wrinkling}
{\sc Popa, T.}, {\sc Zhou, Q.}, {\sc Bradley, D.}, {\sc Kraevoy, V.}, {\sc Fu,
  H.}, {\sc Sheffer, A.}, {\sc and} {\sc Heidrich, W.} 2009.
\newblock Wrinkling captured garments using space-time data-driven deformation.
\newblock In {\em Computer Graphics Forum}. Vol.~28. Wiley Online Library,
  427--435.

\bibitem[\protect\citeauthoryear{Protopsaltou, Luible, Arevalo, and
  Magnenat-Thalmann}{Protopsaltou et~al\mbox{.}}{2002}]{protopsaltou2002body}
{\sc Protopsaltou, D.}, {\sc Luible, C.}, {\sc Arevalo, M.}, {\sc and} {\sc
  Magnenat-Thalmann, N.} 2002.
\newblock {\em A body and garment creation method for an Internet based virtual
  fitting room.}
\newblock Springer.

\bibitem[\protect\citeauthoryear{RedBubble}{RedBubble}{2015}]{tshirtSIG}
{\sc RedBubble}. 2015.
\newblock \url{http://www.redbubble.com}.

\bibitem[\protect\citeauthoryear{Robson, Maharik, Sheffer, and Carr}{Robson
  et~al\mbox{.}}{2011}]{robson2011context}
{\sc Robson, C.}, {\sc Maharik, R.}, {\sc Sheffer, A.}, {\sc and} {\sc Carr,
  N.} 2011.
\newblock Context-aware garment modeling from sketches.
\newblock {\em Computers \& Graphics\/}~{\em 35,\/}~3, 604--613.

\bibitem[\protect\citeauthoryear{Rohmer, Popa, Cani, Hahmann, and
  Sheffer}{Rohmer et~al\mbox{.}}{2010}]{rohmer2010animation}
{\sc Rohmer, D.}, {\sc Popa, T.}, {\sc Cani, M.-P.}, {\sc Hahmann, S.}, {\sc
  and} {\sc Sheffer, A.} 2010.
\newblock Animation wrinkling: augmenting coarse cloth simulations with
  realistic-looking wrinkles.
\newblock In {\em ACM Transactions on Graphics (TOG)}. Vol.~29. ACM, 157.

\bibitem[\protect\citeauthoryear{Saaclothes}{Saaclothes}{2015}]{dress1}
{\sc Saaclothes}. 2015.
\newblock \url{http://www.saaclothes.com}.

\bibitem[\protect\citeauthoryear{Scholz and Magnor}{Scholz and
  Magnor}{2006}]{scholz2006texture}
{\sc Scholz, V.} {\sc and} {\sc Magnor, M.} 2006.
\newblock Texture replacement of garments in monocular video sequences.
\newblock In {\em Proceedings of the 17th Eurographics conference on Rendering
  Techniques}. Eurographics Association, 305--312.

\bibitem[\protect\citeauthoryear{Scholz, Stich, Keckeisen, Wacker, and
  Magnor}{Scholz et~al\mbox{.}}{2005}]{scholz2005garment}
{\sc Scholz, V.}, {\sc Stich, T.}, {\sc Keckeisen, M.}, {\sc Wacker, M.}, {\sc
  and} {\sc Magnor, M.} 2005.
\newblock Garment motion capture using color-coded patterns.
\newblock In {\em Computer Graphics Forum}. Vol.~24. Wiley Online Library,
  439--447.

\bibitem[\protect\citeauthoryear{Seo and Magnenat-Thalmann}{Seo and
  Magnenat-Thalmann}{2003}]{seo2003automatic}
{\sc Seo, H.} {\sc and} {\sc Magnenat-Thalmann, N.} 2003.
\newblock An automatic modeling of human bodies from sizing parameters.
\newblock In {\em Proceedings of the 2003 symposium on Interactive 3D
  graphics}. ACM, 19--26.

\bibitem[\protect\citeauthoryear{Sigal, Mahler, Diaz, McIntosh, Carter,
  Richards, and Hodgins}{Sigal et~al\mbox{.}}{2015}]{sigal2015perceptual}
{\sc Sigal, L.}, {\sc Mahler, M.}, {\sc Diaz, S.}, {\sc McIntosh, K.}, {\sc
  Carter, E.}, {\sc Richards, T.}, {\sc and} {\sc Hodgins, J.} 2015.
\newblock A perceptual control space for garment simulation.
\newblock {\em ACM Transactions on Graphics (TOG)\/}~{\em 34,\/}~4, 117.

\bibitem[\protect\citeauthoryear{Sumner and Popovi{\'c}}{Sumner and
  Popovi{\'c}}{2004}]{sumner2004deformation}
{\sc Sumner, R.~W.} {\sc and} {\sc Popovi{\'c}, J.} 2004.
\newblock Deformation transfer for triangle meshes.
\newblock {\em ACM Transactions on Graphics (TOG)\/}~{\em 23,\/}~3, 399--405.

\bibitem[\protect\citeauthoryear{Tang, Curtis, Yoon, and Manocha}{Tang
  et~al\mbox{.}}{2009}]{tang2009iccd}
{\sc Tang, M.}, {\sc Curtis, S.}, {\sc Yoon, S.-E.}, {\sc and} {\sc Manocha,
  D.} 2009.
\newblock Iccd: Interactive continuous collision detection between deformable
  models using connectivity-based culling.
\newblock {\em Visualization and Computer Graphics, IEEE Transactions
  on\/}~{\em 15,\/}~4, 544--557.

\bibitem[\protect\citeauthoryear{Tanie, Yamane, and Nakamura}{Tanie
  et~al\mbox{.}}{2005}]{tanie2005high}
{\sc Tanie, H.}, {\sc Yamane, K.}, {\sc and} {\sc Nakamura, Y.} 2005.
\newblock High marker density motion capture by retroreflective mesh suit.
\newblock In {\em Robotics and Automation, 2005. ICRA 2005. Proceedings of the
  2005 IEEE International Conference on}. IEEE, 2884--2889.

\bibitem[\protect\citeauthoryear{Taubin}{Taubin}{1995}]{Taubin1995Signal}
{\sc Taubin, G.} 1995.
\newblock A signal processing approach to fair surface design.
\newblock In {\em Proceedings of the 22Nd Annual Conference on Computer
  Graphics and Interactive Techniques}. SIGGRAPH '95. ACM, New York, NY, USA,
  351--358.

\bibitem[\protect\citeauthoryear{Taylor}{Taylor}{2000}]{taylor2000reconstructi%
on}
{\sc Taylor, C.~J.} 2000.
\newblock Reconstruction of articulated objects from point correspondences in a
  single uncalibrated image.
\newblock In {\em Computer Vision and Pattern Recognition, 2000. Proceedings.
  IEEE Conference on}. Vol.~1. IEEE, 677--684.

\bibitem[\protect\citeauthoryear{Thomaszewski, Pabst, and
  Strasser}{Thomaszewski et~al\mbox{.}}{2009}]{thomaszewski2009continuum}
{\sc Thomaszewski, B.}, {\sc Pabst, S.}, {\sc and} {\sc Strasser, W.} 2009.
\newblock Continuum-based strain limiting.
\newblock In {\em Computer Graphics Forum}. Vol.~28. Wiley Online Library,
  569--576.

\bibitem[\protect\citeauthoryear{Turquin, Wither, Boissieux, Cani, and
  Hughes}{Turquin et~al\mbox{.}}{2007}]{turquin2007sketch}
{\sc Turquin, E.}, {\sc Wither, J.}, {\sc Boissieux, L.}, {\sc Cani, M.-P.},
  {\sc and} {\sc Hughes, J.~F.} 2007.
\newblock A sketch-based interface for clothing virtual characters.
\newblock {\em IEEE Computer Graphics and Applications\/}~1, 72--81.

\bibitem[\protect\citeauthoryear{Volino and Magnenat-Thalmann}{Volino and
  Magnenat-Thalmann}{1999}]{volino1999fast}
{\sc Volino, P.} {\sc and} {\sc Magnenat-Thalmann, N.} 1999.
\newblock Fast geometrical wrinkles on animated surfaces.
\newblock In {\em Seventh International Conference in Central Europe on
  Computer Graphics and Visualization (Winter School on Computer Graphics)}.

\bibitem[\protect\citeauthoryear{Wang, Wang, and Yuen}{Wang
  et~al\mbox{.}}{2005}]{wang2005design}
{\sc Wang, C.~C.}, {\sc Wang, Y.}, {\sc and} {\sc Yuen, M.~M.} 2005.
\newblock Design automation for customized apparel products.
\newblock {\em Computer-Aided Design\/}~{\em 37,\/}~7, 675--691.

\bibitem[\protect\citeauthoryear{Wang, O'Brien, and Ramamoorthi}{Wang
  et~al\mbox{.}}{2010}]{wang2010multi}
{\sc Wang, H.}, {\sc O'Brien, J.}, {\sc and} {\sc Ramamoorthi, R.} 2010.
\newblock Multi-resolution isotropic strain limiting.
\newblock In {\em ACM Transactions on Graphics (TOG)}. Vol.~29. ACM, 156.

\bibitem[\protect\citeauthoryear{Wang, O'Brien, and Ramamoorthi}{Wang
  et~al\mbox{.}}{2011}]{wang2011data}
{\sc Wang, H.}, {\sc O'Brien, J.~F.}, {\sc and} {\sc Ramamoorthi, R.} 2011.
\newblock Data-driven elastic models for cloth: modeling and measurement.
\newblock {\em ACM Transactions on Graphics (TOG)\/}~{\em 30,\/}~4, 71.

\bibitem[\protect\citeauthoryear{Weil}{Weil}{1986}]{weil1986synthesis}
{\sc Weil, J.} 1986.
\newblock The synthesis of cloth objects.
\newblock {\em ACM Siggraph Computer Graphics\/}~{\em 20,\/}~4, 49--54.

\bibitem[\protect\citeauthoryear{White, Crane, and Forsyth}{White
  et~al\mbox{.}}{2007}]{white2007capturing}
{\sc White, R.}, {\sc Crane, K.}, {\sc and} {\sc Forsyth, D.~A.} 2007.
\newblock Capturing and animating occluded cloth.
\newblock In {\em ACM Transactions on Graphics (TOG)}. Vol.~26. ACM, 34.

\bibitem[\protect\citeauthoryear{Wu}{Wu}{2011}]{wu2011visualsfm}
{\sc Wu, C.} 2011.
\newblock Visualsfm: A visual structure from motion system.
\newblock {\em URL: http://homes. cs. washington. edu/\~{} ccwu/vsfm\/}~{\em
  9}.

\bibitem[\protect\citeauthoryear{Wu}{Wu}{2013}]{Wu2013Towards}
{\sc Wu, C.} 2013.
\newblock Towards linear-time incremental structure from motion.
\newblock In {\em 3D Vision - 3DV 2013, 2013 International Conference on}.
  127--134.

\bibitem[\protect\citeauthoryear{Xie and Tu}{Xie and
  Tu}{2015}]{xie2015holistically}
{\sc Xie, S.} {\sc and} {\sc Tu, Z.} 2015.
\newblock Holistically-nested edge detection.
\newblock In {\em Proceedings of the IEEE International Conference on Computer
  Vision}. 1395--1403.

\bibitem[\protect\citeauthoryear{Yamaguchi, Kiapour, and Berg}{Yamaguchi
  et~al\mbox{.}}{2013}]{yamaguchi2013paper}
{\sc Yamaguchi, K.}, {\sc Kiapour, M.~H.}, {\sc and} {\sc Berg, T.} 2013.
\newblock Paper doll parsing: Retrieving similar styles to parse clothing
  items.
\newblock In {\em Computer Vision (ICCV), 2013 IEEE International Conference
  on}. IEEE, 3519--3526.

\bibitem[\protect\citeauthoryear{Yang, Yu, Zhou, Du, Davis, and Yang}{Yang
  et~al\mbox{.}}{2014}]{yang2014semantic}
{\sc Yang, Y.}, {\sc Yu, Y.}, {\sc Zhou, Y.}, {\sc Du, S.}, {\sc Davis, J.},
  {\sc and} {\sc Yang, R.} 2014.
\newblock Semantic parametric reshaping of human body models.
\newblock In {\em 3D Vision (3DV), 2014 2nd International Conference on}.
  Vol.~2. IEEE, 41--48.

\bibitem[\protect\citeauthoryear{Ye, Wang, Deng, Yang, and Yang}{Ye
  et~al\mbox{.}}{2014}]{ye2014real}
{\sc Ye, M.}, {\sc Wang, H.}, {\sc Deng, N.}, {\sc Yang, X.}, {\sc and} {\sc
  Yang, R.} 2014.
\newblock Real-time human pose and shape estimation for virtual try-on using a
  single commodity depth camera.
\newblock {\em IEEE transactions on visualization and computer graphics\/}~{\em
  20,\/}~4, 550--559.

\bibitem[\protect\citeauthoryear{Young, Adelstein, and Ellis}{Young
  et~al\mbox{.}}{2007}]{young2007calculus}
{\sc Young, S.}, {\sc Adelstein, B.}, {\sc and} {\sc Ellis, S.} 2007.
\newblock Calculus of nonrigid surfaces for geometry and texture manipulation.
\newblock {\em Visualization and Computer Graphics, IEEE Transactions
  on\/}~{\em 13,\/}~5, 902--913.

\bibitem[\protect\citeauthoryear{Zhao, Chellappa, Phillips, and Rosenfeld}{Zhao
  et~al\mbox{.}}{2003}]{zhao2003face}
{\sc Zhao, W.}, {\sc Chellappa, R.}, {\sc Phillips, P.~J.}, {\sc and} {\sc
  Rosenfeld, A.} 2003.
\newblock Face recognition: A literature survey.
\newblock {\em ACM computing surveys (CSUR)\/}~{\em 35,\/}~4, 399--458.

\bibitem[\protect\citeauthoryear{Zhou, Chen, Fu, Guo, and Tan}{Zhou
  et~al\mbox{.}}{2013}]{zhou2013garment}
{\sc Zhou, B.}, {\sc Chen, X.}, {\sc Fu, Q.}, {\sc Guo, K.}, {\sc and} {\sc
  Tan, P.} 2013.
\newblock Garment modeling from a single image.
\newblock In {\em Computer Graphics Forum}. Vol.~32. Wiley Online Library,
  85--91.

\bibitem[\protect\citeauthoryear{Zhou, Fu, Liu, Cohen-Or, and Han}{Zhou
  et~al\mbox{.}}{2010}]{zhou2010parametric}
{\sc Zhou, S.}, {\sc Fu, H.}, {\sc Liu, L.}, {\sc Cohen-Or, D.}, {\sc and} {\sc
  Han, X.} 2010.
\newblock Parametric reshaping of human bodies in images.
\newblock In {\em ACM Transactions on Graphics (TOG)}. Vol.~29. ACM, 126.

\end{thebibliography}
\end{document}